\pdfoutput=1
\documentclass{article}

\PassOptionsToPackage{numbers,square,comma,sort&compress}{natbib}
\usepackage[eandd,final]{neurips_2026}

\usepackage[T1]{fontenc}
\usepackage[utf8]{inputenc}
\usepackage{microtype}
\usepackage{graphicx}
\usepackage{booktabs}
\usepackage{multirow}
\usepackage{tabularx}
\usepackage{array}
\usepackage{amsmath,amssymb,mathtools}
\usepackage{enumitem}
\usepackage{xspace}
\usepackage{xcolor}
\usepackage{url}
\usepackage{hyperref}
\usepackage{caption}
\usepackage{makecell}
\usepackage{pifont}
\usepackage{adjustbox}
\usepackage{wrapfig}
\usepackage{adjustbox}
\usepackage{multirow}

\hypersetup{hidelinks}

\newcommand{\datasetname}{\textsc{COCOTree}\xspace}

\newcommand{\numImages}{21K\xspace} 
\newcommand{\numMasks}{1.8M\xspace} 

\newcommand{\numLocalLabels}{3.5K\xspace} 

\newcommand{\avgDepth}{3.45\xspace} 

\newcommand{\xsArea}{10^2}
\newcommand{\smallArea}{32^2}
\newcommand{\mediumArea}{96^2}

\newcommand{\projecturl}{\texttt{https://github.com/melonkick3090/COCOTree}}

\newcolumntype{Y}{>{\raggedright\arraybackslash}X}
\newcolumntype{C}[1]{>{\centering\arraybackslash}p{#1}}
\newcolumntype{L}[1]{>{\raggedright\arraybackslash}p{#1}}

\newcommand{\rev}[1]{#1}
\newcommand{\sys}{\datasetname}

\newcommand{\figplaceholder}[2]{%
\begin{figure}[t]
  \centering
  \fbox{\parbox[c][#1][c]{0.94\linewidth}{\centering #2}}
}
\newcommand{\closefigplaceholder}[2]{%
  \caption{#1}
  \label{#2}
\end{figure}
}

\title{\datasetname: A Dataset and Benchmark for Open Tree-Structured Visual Decomposition}

\author{%
  \rev{Junhyub Lee} \\
  Chung-Ang University \\
  \texttt{junhyub3090@cau.ac.kr} \\
  \And
  \rev{Seunghun Chae} \\
  Chung-Ang University \\
  \texttt{ch040602@cau.ac.kr}\\
  \And
  \rev{Hyosu Kim} \\
  Chung-Ang University \\
  \texttt{hskimhello@cau.ac.kr}\\
}

\begin{document}
\maketitle

\begin{abstract}
We formalize and enable the task of \textit{open tree decomposition}, which segments an image into hierarchical trees of visual components with unconstrained granularity and flexibility. Specifically, we provide the foundation benchmark for this new paradigm with the following three key contributions. First, we overcome the prohibitively high cognitive and physical bottlenecks of manual annotation by developing a fully automated generation pipeline that synergizes the semantic reasoning of Large Vision-Language Models (LVLMs) with the precise geometric grounding of SAM 3. Second, leveraging this pipeline, we construct \datasetname, a massive-scale benchmark featuring over 21K images and 1.8M structural nodes. By embracing an open-vocabulary space of over 3.5K unique labels, it successfully captures the long-tail distribution of complex physical assemblies. Notably, rigorous human evaluation confirms our generated annotations demonstrate strong alignment with human structural judgment. Third, we establish a standardized evaluation protocol by proposing the Open Tree Quality (OTQ) metric, which jointly assesses mask precision, label accuracy, and structural consistency. We release our dataset and benchmark code at \projecturl.
\end{abstract}

\section{Introduction}\label{sec:introduction}
Image segmentation has long served as a fundamental pillar of visual recognition. It partitions an image into either semantic category regions, distinct object instances, or panoptic outputs at the pixel level~\citep{long2015fcn,he2017maskrcnn,kirillov2019panoptic}. In particular, recent advances in image segmentation have enabled the segmentation of images into extremely fine-grained units, e.g., constituent parts of objects~\citep{myersdean2024spin}, providing granular comprehension of complex physical environments. 

However, existing methods typically provide a single-level abstraction of these fine-grained components, entirely ignoring the hierarchical dependencies inherent in real-world assemblies. This lack of structural information not only hinders compositional reasoning over visual scenes but also causes severe functional ambiguity: an isolated "handle" segment provides no structural clues as to whether it can be used to open a door, hold a cup, or turn a mechanical valve. These problems pose a critical bottleneck for autonomous systems, such as embodied AI agents, which require complex physical manipulation rather than mere visual perception.

In response to the need for hierarchical image understanding, hierarchical datasets~\citep{pascalpart,degeus2021partaware,he2022partimagenet,ramanathan2023paco,li2022panopticpartformer} have been introduced. These datasets capture compositional relationships but with severe limitations in granularity and flexibility. First, they primarily confine their structural depth to single-step object-part dependencies, rarely extending to the deeper sub-part hierarchies. This shallow abstraction thus loses the deep, critical structural chain of objects (e.g., tracing from a car, to its door, but not down to the specific door handle). Furthermore, these structural annotations are strictly constrained by closed-set vocabularies. Such a rigid taxonomy-based annotation might be effective for object recognition, but is entirely inadequate for modeling the unconstrained compositional structures inherent to the real world. That is, these datasets ignore any novel, long-tail components or structural configurations falling outside the taxonomies, although they are clearly present in the images.

In this paper, we facilitate the development of algorithms for hierarchical analysis with unconstrained granularity and flexibility, a task which we call \textit{open tree decomposition}. Specifically, we provide a large-scale dataset in which images are represented as hierarchical trees of visible components, called \textit{open trees}. Each node in the tree encapsulates a single instance mask paired with its corresponding semantic label. In particular, the tree's hierarchical structure and each node's label are derived from the unconstrained visual reality of the image, rather than forced into a predefined template. 

However, constructing such highly granular annotations through human labor is prohibitively challenging at scale. Even worse, deriving unconstrained structural and linguistic annotations based on the given visual evidence exponentially amplifies the cognitive burden. To overcome the fundamental limitations of manual annotation, we develop a fully automated annotation pipeline that recursively leverages recent Large Vision-Language Models (LVLMs) and SAM 3~\cite{carion2026sam3}. Starting from the entire scene, it prompts LVLMs for a semantic decomposition of the current visual region. These semantic proposals are then spatially grounded into precise instance masks by SAM 3. Each generated mask is subsequently cropped and fed back into the LVLM as a new parent node, repeating this cycle until the model determines no further meaningful subdivisions exist.

\begin{figure}[t]
    \centering
    \includegraphics[width=\textwidth]{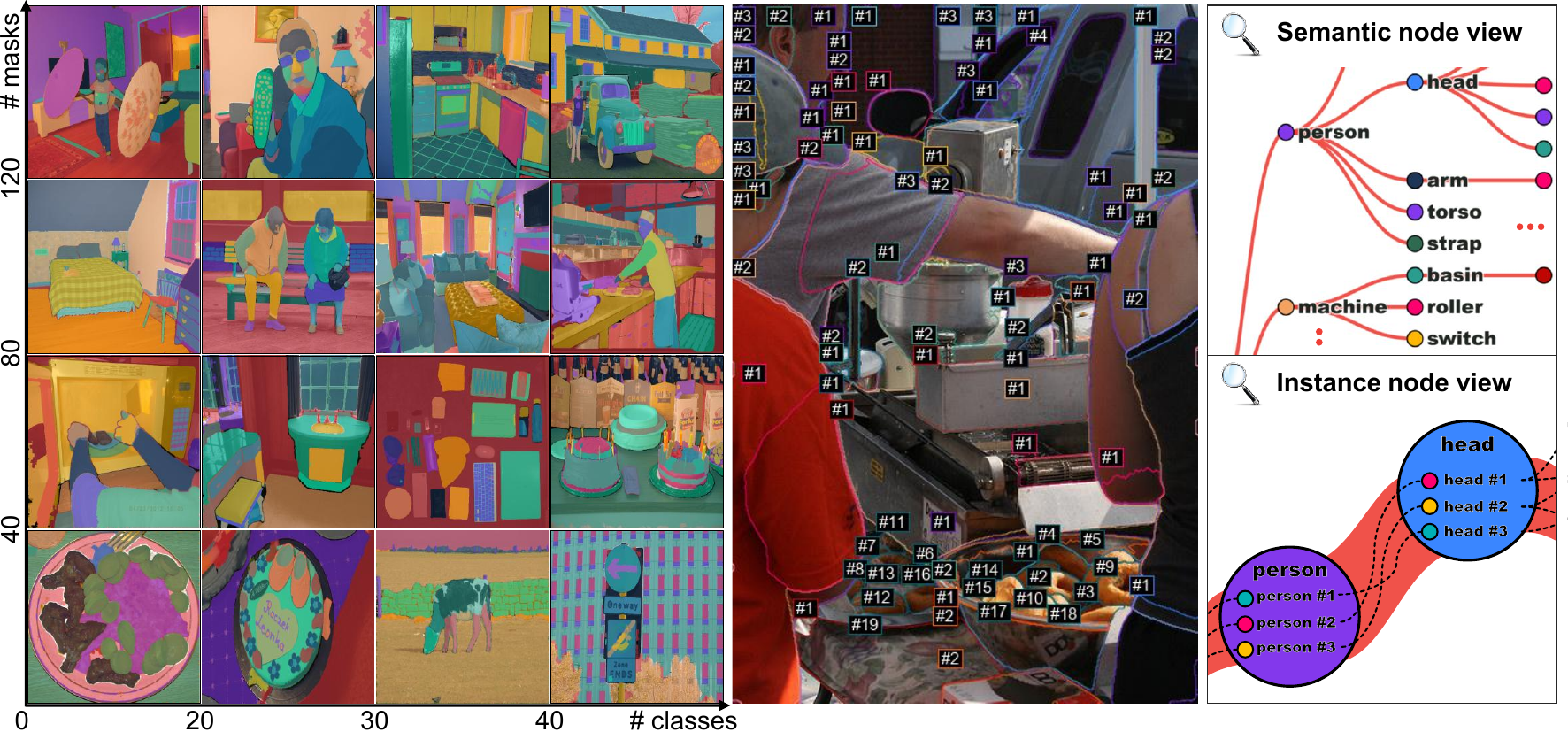}
    \vspace{-0.2cm}
    \caption{Overview of \datasetname. Left: \datasetname provides dense open-tree annotations over COCO images.
    Middle: each image is decomposed into visible components grounded by instance masks.
    Right: the same annotation can be viewed as a semantic-node tree, which groups repeated masks under a shared local label, and as an instance-node tree, where each mask becomes a node with an explicit visual parent.
    \vspace{-0.5cm}}
    \label{fig:teaser}
\end{figure}

Using this automated pipeline, we construct \datasetname. Originating from the COCO dataset~\cite{lin2014microsoft}, \datasetname benefits from a high density of complex, interacting objects within everyday environments like other COCO-based datasets~\citep{gupta2019lvis, deng2024coconut, ramanathan2023paco, degeus2021partaware}, while simultaneously providing unprecedented hierarchical granularity and flexibility (see examples in Figure~\ref{fig:teaser}). It features over 21K images and 1.8M total instance masks (i.e., nodes), achieving an incredibly dense annotation rate of 85.7 masks per image and an average tree depth of \avgDepth. Furthermore, it catalogs over 3.5K unique open-vocabulary labels, successfully capturing the long-tail distribution of real-world physical components. It should be noted that we ensured the practical utility of these automated annotations by assessing their quality via extensive human evaluation, which confirms strong alignment with human judgment regarding both hierarchical structures and open-vocabulary labeling. \rev{\datasetname is designed as a scalable silver reference for open tree decomposition, constructed under a consistent visible-evidence annotation policy. Under this framing, \datasetname serves as a benchmark for open tree decomposition tasks with a metric called \textit{Open Tree Quality (OTQ)}.} This is a PQ (Panoptic Quality~\citep{kirillov2019panoptic})-like metric that jointly evaluates mask quality, label quality, and structural consistency.

\section{Related Work}
\label{Sec:relatedworks}

\subsection{Hierarchical Segmentation}
Hierarchical segmentation datasets have extended image segmentation beyond a single level of abstraction by annotating visual structures. \rev{Specifically, numerous benchmarks capture the compositional relationships between parent objects and their part components~\cite{pascalpart,he2022partimagenet,li2024pinpp,ramanathan2023paco,meletis2020panopticparts}.} Recently, this hierarchical granularity has been further increased to capture part-subpart relationships in natural images~\cite{zhou2019semantic, myersdean2024spin}. Despite these advancements, existing paradigms force visual representations into rigid, predefined templates. By limiting structural hierarchies to a shallow, two- or three-tier depth and restricting labels to a closed vocabulary, these datasets strip away the unconstrained structural context of the physical world. This shallow abstraction creates a fundamental data bottleneck for complex compositional reasoning.
In addition, diverse metrics have been introduced as evaluation protocols for hierarchical segmentation. For instance, PartPQ (part panoptic quality)~\cite{degeus2021partaware} extends standard panoptic quality to assess segmentation performance across a strict two-level object-part hierarchy, while HPQ (hierarchical panoptic quality)~\cite{tang2023visual} broadens this to evaluate fixed, multi-tier semantic hierarchies. Despite their utility, both metrics systematically fail when applied to unconstrained granularity, open-vocabulary structures.

Consequently, hierarchical segmentation models have traditionally mirrored these structural constraints. While they support the prediction of basic object-part relations~\cite{li2022panopticpartformer} or even more granular hierarchies~\cite{li2022hssn, wang2023hipie,li2023semanticsam,tang2023visual}, they inherently operate within predefined semantic taxonomies or structural depths.

\subsection{Hierarchical Vision-Language Segmentation}
With recent advances in LVLMs~\cite{liu2023visualinstructiontuning}, the paradigm of image segmentation has increasingly shifted towards open-vocabulary, instruction-driven parsing. Recent frameworks seamlessly couple the semantic reasoning of LVLMs with dense segmentation tasks~\cite{lai2024lisa, ren2024pixellm, wang2024llmseg, rasheed2024glamm}. In particular, GLaMM~\cite{rasheed2024glamm} demonstrates that automated annotation pipelines powered by LVLMs can successfully drive the generation of large-scale segmentation datasets. In addition, the visual reasoning capabilities of LVLMs have been integrated into dense segmentation pipelines~\cite{wang2024llmseg,chen2024sam4mllm,wang2024segllm,zhang2025think2segrs}, typically by prompting geometric foundation models like SAM~\cite{kirillov2023segment} and SAM 3~\cite{carion2026sam3}. However, these systems inherently treat segmentation as a static, single-level task, resulting in a lack of the structural awareness necessary to parse complex, multi-level compositional assemblies.

To bridge this gap, several recent methods have moved from flat open-vocabulary segmentation toward structured, hierarchical prediction. Language-conditioned segmentors now produce masks at varying semantic levels, leveraging the reasoning capabilities of LVLMs to generate grounded hierarchical outputs~\cite{wang2023hipie,li2023semanticsam,tang2023visual,lai2024lisa,ren2024pixellm,wang2024llmseg,rasheed2024glamm}. HALLUMI~\cite{myersdean2025hallumi} generates hierarchical segmentation masks via autoregressive language modeling to capture object-part-subpart granularity under predefined structural taxonomies.  

\section{\datasetname: Task, Construction, and Dataset}
\label{sec_cocotree}
For a comprehensive understanding of \datasetname, a dataset and benchmark for open tree decomposition, we first formulate the open tree decomposition task, detail the automated recursive annotation pipeline for constructing a large-scale hierarchical dataset, and summarize the scale and structure of the resulting dataset.

\subsection{Open Tree Decomposition Task Formulation}\label{sec_opentree}
Given an input image $I \in \mathbb{R}^{H \times W \times 3}$, traditional hierarchical segmentation tasks typically map pixels to a predefined set of categories $C$ with a strictly bounded hierarchical depth. In contrast, we formulate open tree decomposition as the task of parsing the image $I$ into an unconstrained hierarchical tree, called an \textit{open tree}, directly derived from the visual evidence of $I$. The open tree is structured as $T = (V, E)$:
\begin{itemize}[leftmargin=*, itemsep=2pt, topsep=2pt]
    \item \textit{Nodes ($V$, visible components).} Each node $v_i \in V$ represents a distinct visual component in the image $I$ and is defined as a tuple $v_i = (m_i, l_i)$. Here, $m_i \in \{0, 1\}^{H \times W}$ is a binary instance mask spatially grounding the component, and $l_i \in \mathcal{L}$ is its corresponding semantic label. Unlike prior work, $\mathcal{L}$ represents an open-vocabulary linguistic space, allowing $l_i$ to capture any long-tail or unconstrained text description.

    \item\textit{Edges ($E$, structural hierarchy).} The edge set $E$ captures the compositional relationships between visible components. A directed edge $e_{i,j} \in E$ (from parent $v_i$ to child $v_j$) exists if and only if $v_j$ is a constituent sub-part visually encompassed by $v_i$ in the image $I$.
\end{itemize}
\rev{The fundamental distinction of $T$ is its structural flexibility without a predefined maximum depth. The tree originates from a universal root node $r$ (representing the entire scene $I$) and recursively branches based on the visually supported component structure of the image, terminating at leaf nodes where no further meaningful sub-divisions exist. An image may allow multiple valid open trees with different decomposition granularities and structures. In \datasetname, we instantiate one such decomposition under a consistent visible-evidence annotation policy.}

\begin{figure}[t]
    \centering
    \includegraphics[width=\textwidth]{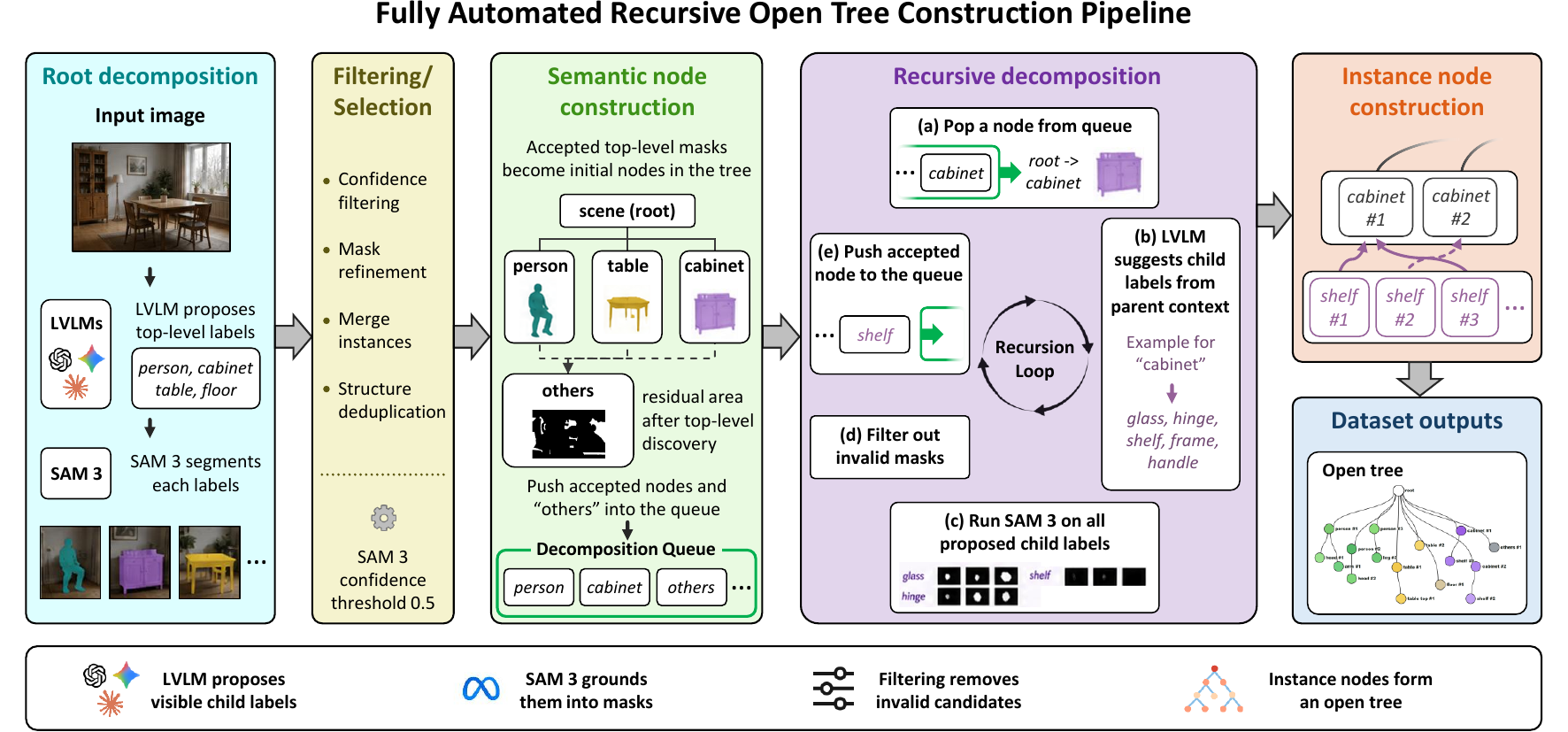}
    \vspace{-0.2cm}
    \caption{Fully automated open tree construction pipeline.}
    \label{fig:pipeline}
\end{figure}

\subsection{Fully Automated Open Tree Annotation Pipeline} \label{sec_pipeline}
Open tree annotation is a recursive generation process that derives labels, masks, and structural relationships directly from visual evidence. However, executing this unconstrained decomposition through manual human annotation presents severe scalability and consistency bottlenecks. We therefore automate open tree annotation with a recursive pipeline that combines the unconstrained semantic reasoning of LVLMs with the highly dense, precise localization of vision foundation models (SAM 3). In particular, to ensure computational scalability, this recursive decomposition is driven at the semantic level rather than the instance level in the following steps (see Figure~\ref{fig:pipeline}).

\paragraph{Root decomposition.}
The generation process initializes by deconstructing the highest-level visual context (the root node). First, the LVLM analyzes the unconstrained image to propose common-noun labels for the major constituent elements in the scene. The proposal prompt relies strictly on visible evidence, filtering out materials, abstract attributes, and near-synonyms to prevent hallucination (see Appendix~\ref{app:prompts}). Second, these semantic proposals are passed to the vision foundation model, which grounds them into precise spatial masks, isolating primary objects, distinct scene regions, and broad contextual background elements (i.e., 'stuff'). A single proposed label may yield multiple grounded masks corresponding to distinct visible instances; these are structurally grouped to form a semantic node. The geometric union of these grouped masks defines the visual context for subsequent recursive decomposition, while the individual instance masks are retained for the final open tree materialization. Finally, these foundational semantic nodes are connected via directed edges to the root node. Note that this initial pipeline also records an \texttt{others} region for uncovered pixels used for further decomposition.

\paragraph{Recursive decomposition.}
Once initialized, all newly generated semantic nodes are enqueued to drive the recursive expansion of the tree. For each node within the queue, the pipeline extracts a strictly isolated visual representation: the original image is tightly cropped to the spatial extents of the node’s grouped mask, and all external background pixels are suppressed. \rev{Crops from previous recursive steps are not reused to generate deeper crops, avoiding cumulative resampling across recursion.} This localized visual crop is then fed back into the LVLM with its corresponding semantic label and hierarchical structure, re-initiating the semantic proposal phase to uncover sub-components. \rev{If a proposed child cannot be grounded by SAM~3 with an accepted mask, the corresponding node is omitted and that branch is not expanded further, while other successfully grounded sibling branches continue independently.} This iterative loop dynamically expands the tree, treating every accepted part as a new parent until no further visually supported sub-components can be proposed and grounded.

\paragraph{Evidence-based filtering.}
During the decomposition process, the pipeline retains semantic nodes only if they are supported by sufficient grounding evidence. To achieve this, we employ a scale-adaptive filtering strategy for child mask extraction. Specifically, the confidence threshold of the vision foundation model is dynamically adjusted based on the parent node's spatial footprint: if the parent mask occupies less than $5\%$ of the full image area, the threshold is relaxed to $0.4$ to ensure the successful capture of fine-grained sub-parts; otherwise, a stricter threshold of $0.5$ is applied to prevent the extraction of low-confidence noise in larger regions. At the scene level, we reject top-level candidates whose masks cover more than $70\%$ of the full image to prevent near-global concepts such as ``scene'' or ``image'' from becoming ordinary nodes. We also reject child nodes whose masks cover more than $90\%$ of their immediate parent mask. For sibling cleanup, we resolve duplicate component proposals by evaluating their geometric intersection. If the spatial overlap between two sibling masks exceeds $90\%$, they are merged into a single representation to eliminate redundant structural branches. \rev{These thresholds are empirically chosen heuristics for the \datasetname construction pipeline. A sensitivity analysis supports the selected SAM~3 confidence settings, which may require recalibration for other image distributions (Appendix~\ref{app:confidence_threshold}).}

\paragraph{Instance-level open tree conversion.} After recursive semantic decomposition is complete, the intermediate tree of semantic nodes is materialized into an instance-level hierarchy. First, the grouped masks within each semantic node are separated. Each individual mask is instantiated as a distinct node paired with its semantic label. Next, to establish instance-level structural dependencies, lower-level instance masks are assigned to their corresponding parent masks based on spatial containment and overlap heuristics. In cases of ambiguity where a child spatially intersects multiple potential parents, it is deterministically routed to the parent demonstrating the strongest containment evidence. Conversely, child nodes lacking a reliable parent are rejected as noise.

\subsection{\datasetname Analysis}
\label{sec_dataset}
Leveraging the aforementioned automated pipeline across the COCO dataset, we constructed \datasetname. Before establishing its utility as an evaluation benchmark, we validate \datasetname as a standalone reference dataset as follows\footnote{We use four mask-size bins for dataset analysis and flat-mask compatibility:
 \(\mathrm{XS}: 0<A<\xsArea\),
 \(\mathrm{S}^{\star}: \xsArea\le A<\smallArea\),
 \(\mathrm{M}: \smallArea\le A<\mediumArea\), and
 \(\mathrm{L}: A\ge\mediumArea\)}.

\begin{table*}[t]
\centering
\small
\renewcommand{\arraystretch}{1.0}

\begin{minipage}[t]{0.60\textwidth}
\vspace{0pt}
\centering
\caption{Dataset scale and density. M/img denotes masks per image. Max$D$ denotes maximum hierarchy depth.}
\label{tab:dataset_comparison}
\vspace{0.4em}
\setlength{\tabcolsep}{2.4pt}
\begin{tabular*}{\linewidth}{@{\extracolsep{\fill}}lccccc@{}}
\toprule
Dataset & Images & Masks & M/img & Classes & Max$D$ \\
\midrule
COCO-17~\cite{lin2014microsoft}       & 118K & 860K & 7.3  & 80   & 1 \\
COCO-Stuff~\cite{caesar2018cocostuff} & 118K & 1.0M & 8.5  & 91   & 1 \\
COCONut~\cite{deng2024coconut}        & 358K & 4.7M & 13.1 & 133  & 1 \\
COCO-ReM~\cite{singh2024benchmarking} & 118K & 1.1M & 9.3  & 80   & 1 \\
LVIS~\cite{gupta2019lvis}             & 100K & 1.3M & 13.0 & 1.2K & 1 \\
\midrule
PACO~\cite{ramanathan2023paco}         & 77K  & 901K & 11.7 & 531  & 2 \\
CPP~\cite{degeus2021partaware}         & 3.5K & 156K & 44.6 & 28   & 2 \\
PartImageNet~\cite{he2022partimagenet} & 24K  & 136K & 5.7  & 198  & 2 \\
\rev{PartImageNet++~\cite{li2024pinpp}}      & 100K & 406K & 4.1  & 4.3K & 2 \\
SPIN~\cite{myersdean2024spin}          & 10K  & 146K & 14.6 & 254  & 3 \\
ADE20K~\cite{zhou2019semantic}         & 20K  & 270K & 13.5 & 150  & 3 \\
\textbf{\datasetname} & \textbf{\numImages} & \textbf{\numMasks} & \textbf{85.7} & \textbf{\numLocalLabels} & \textbf{Unconstrained} \\ 
\bottomrule
\end{tabular*}
\end{minipage}
\hfill
\begin{minipage}[t]{0.36\textwidth}
\vspace{0pt}
\centering
\caption{Node depth distribution by size bin. Columns sum to 100\%.}
\label{tab:depth_size_stats}
\vspace{0.4em}
\setlength{\tabcolsep}{2.0pt}
\begin{tabular*}{\linewidth}{@{\extracolsep{\fill}}cccccc@{}}
\toprule
Depth & All & XS & $S^\star$ & M & L \\
\midrule
1       & 27.7 & 7.0  & 21.3 & 36.9 & 69.8 \\
2       & 42.3 & 36.6 & 45.0 & 45.9 & 26.1 \\
3       & 23.9 & 39.1 & 27.6 & 15.6 & 3.9  \\
$\geq4$ & 6.1  & 17.3 & 6.1  & 1.6  & 0.2  \\
\bottomrule
\end{tabular*}
\end{minipage}
\vspace{0.35em}
\end{table*}

\paragraph{Scale and structure.}
As summarized in Table~\ref{tab:dataset_comparison}, \datasetname achieves an unprecedented annotation density, averaging 85.7 masks per image. This massive scale dwarfs traditional flat-hierarchy baselines like LVIS (13.0 masks/img) and COCONut (13.1 masks/img), and nearly doubles the density of the most granular hierarchical baseline, CPP (44.6 masks/img). Structurally, while existing hierarchical datasets are bounded to maximum depths of 2 or 3, our automated pipeline expands to an unconstrained depth. \rev{Notably, this unconstrained depth allows different branches to terminate at different levels based on visible evidence.} While 69.8\% of Large (L) masks reside at depth 1 (representing macro-objects), the tree successfully isolates fine-grained sub-parts, with 56.4\% of Extra Small (XS) masks securely grounded at depths of 3 or greater (see Table~\ref{tab:depth_size_stats}).

\paragraph{Mask compatibility.}
To verify the geometric quality of our automated annotations, Table~\ref{tab:mask_compatibility} evaluates the mask compatibility of \datasetname against established flat-hierarchy datasets. Overall, our generated masks exhibit high alignment with manual human annotations, achieving median IoUs ranging from $0.73$ (against LVIS) to $0.90$ (against COCO-ReM). In particular, it recovers a high proportion of Large ($L$) and Medium ($M$) reference masks (e.g., $AR_L$ of $0.90$ against COCO-ReM). However, recall drops for XS masks. Such tiny regions are difficult for LVLMs to propose from visual evidence and for the visual foundation model to ground reliably. \rev{The reference datasets primarily annotate objects, while \datasetname also captures fine-grained parts and subparts. Additional analysis shows that only $16.4$--$34.7\%$ of their XS masks have corresponding matches in \datasetname, yet the matched pairs achieve mean IoUs of $0.706$--$0.800$ (Appendix~\ref{app:small_mask_overlap}). These results highlight differences in annotation coverage, indicating that low XS recall alone does not necessarily reflect poor mask quality.}

\paragraph{Human validation.}
Our human validation study comprised 20 reviewers who each analyzed 50 randomly selected images paired with their corresponding open trees. Utilizing a custom web interface, the reviewers qualitatively assessed the accuracy and quality of every semantic node and overall tree structures (the interface UI and full questionnaire are detailed in Appendix~\ref{app:review_web}). As reported in Table~\ref{tab_human_validation_response_distribution}, the generated open trees exhibit high quality at the node level, achieving $\geq91\%$ Good ratings for semantic matching (Q1), instance separation (Q4), child validity (Q6), and leaf appropriateness (Q8). \rev{Boundary quality also remains broadly stable at deeper levels, with Good-or-Minor ratings of $96.06\%$ at depth 3 and $95.43\%$ at depth $\geq4$ (Appendix~\ref{app:depth_validation}). At the tree level, full-tree consistency (T1) and missing nodes (T2) received Minor ratings of $44.3\%$ and $41.6\%$, respectively, while Major ratings remained at $1.0\%$ and $7.7\%$. For T2, Minor responses included comments such as ``A small visible portion of a white chair left unannotated'' and ``Some fine lettering and markings are only partially captured,'' indicating that such ratings do not necessarily imply missing critical components. These results confirm that our fully automated pipeline generates structurally coherent, high-fidelity hierarchical annotations that align closely with human perception.}

\begin{table*}[t]
\caption{Mask compatibility with existing flat-hierarchy segmentation annotations. Size columns report AR by mask-size bin.}
\label{tab:mask_compatibility}
\centering
\small
\setlength{\tabcolsep}{4pt}
\renewcommand{\arraystretch}{1.0}
\begin{tabular*}{\textwidth}{@{\extracolsep{\fill}}lccccccccc@{}}
\toprule
& \multicolumn{2}{c}{IoU$\uparrow$} & \multicolumn{7}{c}{AR$\uparrow$} \\
\cmidrule(lr){2-3}
\cmidrule(lr){4-10}
Reference & Mean & Median & AR & AR@50 & AR@75 & AR$_{XS}$ & AR$_{S^\star}$ & AR$_M$ & AR$_L$ \\
\midrule
COCO-17    & 0.69 & 0.81 & 0.56 & 0.80 & 0.62 & 0.13 & 0.45 & 0.64 & 0.82 \\
COCONut    & 0.69 & 0.83 & 0.58 & 0.80 & 0.64 & 0.14 & 0.42 & 0.69 & 0.88 \\
COCO-ReM   & 0.69 & 0.90 & 0.63 & 0.74 & 0.67 & 0.15 & 0.52 & 0.80 & 0.90 \\
LVIS       & 0.54 & 0.73 & 0.45 & 0.60 & 0.49 & 0.08 & 0.40 & 0.71 & 0.87 \\
\bottomrule
\vspace{-7mm}
\end{tabular*}
\end{table*}

\begin{table}[t]
\caption{Human validation response distributions. Values are percentages with standard deviations.}
\label{tab_human_validation_response_distribution}
\centering
\small
\setlength{\tabcolsep}{4pt}
\begin{tabular*}{\linewidth}{@{\extracolsep{\fill}}l l c c c c c@{}}
\toprule
Q & Axis & Good$\uparrow$ & Minor & Moderate & Major$\downarrow$ & Not clear \\
\midrule
\multicolumn{7}{l}{\textit{Node-level questions}} \\
Q1 & Label-mask match  & 92.9 $\pm$ 2.1  & -- & -- & 5.2 $\pm$ 1.5 & 2.0 $\pm$ 1.7 \\
Q2 & Target coverage     & 71.7 $\pm$ 4.7  & 18.5 $\pm$ 4.1 & -- & 9.6 $\pm$ 3.6 & 0.1 $\pm$ 0.1 \\
Q3 & Extra instance      & 88.3 $\pm$ 3.2  & 9.1 $\pm$ 2.3  & -- & 2.5 $\pm$ 2.0 & 0.2 $\pm$ 0.2 \\
Q4 & Instance separation & 91.4 $\pm$ 7.3  & 5.8 $\pm$ 4.9 & 2.4 $\pm$ 2.5 & 0.3 $\pm$ 0.6 & 0.1 $\pm$ 0.3 \\
Q5 & Boundary quality    & 79.2 $\pm$ 11.3 & 16.8 $\pm$ 10.2 & 3.5 $\pm$ 2.2 & 0.4 $\pm$ 0.7 & 0.1 $\pm$ 0.2 \\
Q6 & Child validity      & 93.3 $\pm$ 5.9  & 4.9 $\pm$ 4.7 & -- & 1.3 $\pm$ 1.9 & 0.5 $\pm$ 1.0 \\
Q7 & Child missing       & 91.8 $\pm$ 9.1  & 7.3 $\pm$ 8.2 & -- & 0.6 $\pm$ 0.9 & 0.3 $\pm$ 0.7 \\
Q8 & Leaf appropriateness& 98.4 $\pm$ 1.5  & -- & -- & 1.5 $\pm$ 1.5 & 0.1 $\pm$ 0.1 \\
\midrule
\multicolumn{7}{l}{\textit{Tree-level questions}} \\
T1 & Tree consistency       & 47.5 $\pm$ 23.7 & 44.3 $\pm$ 19.9 & 7.1 $\pm$ 6.8 & 1.0 $\pm$ 2.0 & 0.1 $\pm$ 0.4 \\
T2 & \rev{Missing nodes} & 50.5 $\pm$ 16.9 & 41.6 $\pm$ 13.9 & -- & 7.7 $\pm$ 5.2 & 0.2 $\pm$ 0.6 \\
\bottomrule
\vspace{-7mm}
\end{tabular*}
\end{table}

\section{Open Tree Quality and Benchmarking}
\label{sec_metrics}

\subsection{Metric Design}
\label{sec_metric_design}

\sys requires a metric for image-specific visible trees with open-vocabulary labels. Such outputs are difficult to evaluate because correct masks alone do not guarantee correct open tree predictions: a region may be mislabeled, assigned to the wrong parent, duplicated, missed, or stopped at the wrong decomposition level. We introduce Open Tree Quality (OTQ), a PQ-style metric for open tree decomposition. OTQ evaluates this target in three parts. First, predicted and reference nodes are globally matched by mask IoU. Second, each true positive node is scored by mask IoU and open-label similarity. Third, matched nodes are checked for structural consistency through their visual parents. The final score combines branch quality with the mean quality of matched mask-nodes. \rev{OTQ measures agreement with the released silver reference and does not assume a unique valid decomposition for each image. Its matched-node skeleton accommodates some differences in intermediate structure, while unmatched nodes and structural inconsistencies remain penalized.}

\subsection{Node Matching and Matched-Node Quality}
\label{sec_node_matching_quality}

Let \(G=(V,E)\) be the reference tree and \(P=(\hat V,\hat E)\) be the predicted tree. Each reference node \(v\in V\) has an instance mask \(m_v\), a local label \(\ell_v\), and a parent \(\mathrm{pa}(v)\). Each predicted node \(\hat v\in \hat V\) similarly has \(\hat M_{\hat v}\), \(\hat \ell_{\hat v}\), and \(\mathrm{pa}(\hat v)\).

A one-to-one node assignment \(\mathcal A\) between predicted and reference nodes is obtained by maximum-weight bipartite matching over \(\mathrm{IoU}({\hat v,v})\). 

A matched pair is accepted as a \(TP_{\mathrm{node}}\) when its IoU is at least the node threshold \(\tau_{\mathrm{node}}=0.5\):
\begin{equation}
TP_{\mathrm{node}} = \{(\hat v,v)\in\mathcal A \mid \mathrm{IoU}({\hat v,v})\geq \tau_{\mathrm{node}}\}.
\end{equation}

All predicted nodes not included in \(TP_{\mathrm{node}}\) are \(FP_{\mathrm{node}}\), and all reference nodes not included in \(TP_{\mathrm{node}}\) are \(FN_{\mathrm{node}}\).
\rev{Analysis of IoU distributions against human-annotated references supports $\tau_{\mathrm{node}}=0.5$ as a reasonable matching threshold (Appendix~\ref{app:iou_analysis}).}

For each true-positive node match \((\hat v,v)\), mask quality (MQ) and label quality (LQ) are defined as follows:
\begin{equation}
MQ(\hat v,v)=\mathrm{IoU}({\hat v,v}), \quad
LQ_m(\hat v,v)=\mathrm{Sim}_{m}(\hat \ell_{\hat v},\ell_v),
\end{equation}
where \(m\) denotes the label-similarity protocol, including strict matching,  WordNet/OEWN-based similarity~\cite{miller1995wordnet,mccrae2020english}, BERT/SBERT-based similarity~\cite{devlin2019bert,reimers2019sentencebert},  or another released backend such as CLIP~\cite{radford2021learning}.

The matched-node quality and the mean quality over matched nodes are:
\begin{equation}
NQ_m(\hat v,v)
=
MQ(\hat v,v)\cdot LQ_m(\hat v,v), \quad 
\mathrm{meanNQ}_m
=
\frac{1}{|TP_{\mathrm{node}}|}
\sum_{(\hat v,v)\in TP_{\mathrm{node}}}
NQ_m(\hat v,v).
\label{eq_meannq}
\end{equation}
If there are no true-positive node matches, we set \(\mathrm{meanNQ}_m=0\), and the final OTQ score is zero.

For diagnostics, we also report the average matched-node mask and label terms separately:
\begin{align}
MQ &=
\frac{1}{|TP_{\mathrm{node}}|}
\sum_{(\hat v,v)\in TP_{\mathrm{node}}}
IoU({\hat v,v}), \quad
LQ_m &=
\frac{1}{|TP_{\mathrm{node}}|}
\sum_{(\hat v,v)\in TP_{\mathrm{node}}}
\mathrm{Sim}_{m}(\hat \ell_{\hat v},\ell_v).
\end{align}

\subsection{Branch Quality and Open Tree Quality}
\label{sec_tree_quality}

Tree quality evaluates whether matched nodes preserve the visual organization of the reference tree.
Exact edge matching can be overly strict for open tree decomposition because a prediction may insert or skip an intermediate node while preserving the larger visual ancestor. We therefore compare nearest matched common parents on the matched mask-node skeleton.

Let \(r_G\) and \(r_P\) denote the artificial roots of the reference tree \(G\) and the predicted tree \(P\), respectively.
We form a TP-matched skeleton for each tree by keeping only the true-positive matched nodes and the artificial root. 
For parent assignment, we climb the original semantic-label tree one level at a time and attach the mask-node to the highest-IoU mask under the first ancestor label with positive overlap. 

Let \(G^{TP}\) and \(P^{TP}\) denote the resulting reference and prediction skeletons.

For matched masks $m_i$ and $m_j$, we define their nearest matched common parents as

\begin{equation}
p_G(m_i,m_j)=\mathrm{LCA}_{G^{TP}}(v_i,v_j),
\qquad
p_P(m_i,m_j)=\mathrm{LCA}_{P^{TP}}(\hat v_i,\hat v_j).
\end{equation}

A pair is branch-consistent when the reference-side nearest matched common parent maps to the prediction-side nearest matched common parent:
\begin{equation}
C(m_i,m_j)
=
\mathbf{1}\!\left[
\!\left(p_G(m_i,m_j)\right)
=
p_P(m_i,m_j)
\right].
\end{equation}

Let
\begin{equation}
\mathcal P_{TP}
=
\{\{m_i,m_j\}\mid m_i,m_j\in TP_{\mathrm{node}},\, i<j\}
\end{equation}
be all unordered pairs of true-positive mask-node matches.
The branch-pair accuracy is
\begin{equation}
BQ
=
\begin{cases}
1, & |\mathcal P_{TP}|=0,\\[3pt]
\dfrac{1}{|\mathcal P_{TP}|}
\sum_{\{m_i,m_j\}\in\mathcal P_{TP}}
C(m_i,m_j),
& |\mathcal P_{TP}|>0.
\end{cases}
\label{eq_branch_pair_accuracy}
\end{equation}

Tree quality applies a PQ-style recovery penalty to the branch-pair accuracy:
\begin{equation}
TQ
=
BQ
\cdot
\frac{
|TP_{\mathrm{node}}|
}{
|TP_{\mathrm{node}}|
+\frac12|FP_{\mathrm{node}}|
+\frac12|FN_{\mathrm{node}}|
}.
\label{eq_tree_quality}
\end{equation}
Thus, BQ measures whether recovered mask-nodes preserve visual organization, while the denominator penalizes missing and spurious mask-nodes.

Finally, Open Tree Quality combines tree quality with matched-node quality:
\begin{equation}
OTQ_m
=
TQ\cdot \mathrm{meanNQ}_m .
\label{eq_otq}
\end{equation}
\rev{OTQ extends PQ-style evaluation by combining mask-based node matching, open-label similarity, and tree structure similarity.}
\subsection{OTQ Evaluation with Controlled Degradations}
\label{sec_metric_audit}

\rev{
We evaluate OTQ using controlled degradations of \datasetname annotations on the same 1K-image subset used for human validation in Section~\ref{sec_dataset}. These experiments cover geometric, structural, and semantic changes relevant to open tree decomposition.
Each modified tree is evaluated against its original annotation. Table~\ref{tab_metric_audit} reports the effects of these degradations (see Appendix~\ref{app:otq_additional_results} for additional settings).

Mask erosion and dilation reduce MQ and meanNQ, with erosion also lowering TQ. Parent rewiring reduces TQ while preserving meanNQ, MQ, and LQ at 1.000. Similarly, node removal lowers TQ despite the unchanged quality of the remaining matched nodes. Label rephrasing maintains a high OTQ score (0.982), whereas label rewiring reduces it to 0.842 through LQ. These results demonstrate that OTQ provides interpretable component-level diagnostics for geometric, structural, and semantic changes, complementing the overall hierarchical evaluation of HPQ.
}

\begin{table}[t]
\caption{\rev{Evaluation of OTQ under controlled degradations of \datasetname annotations. All degraded variants use a 50\% keep ratio, which controls mask area changes or the proportion of nodes, parent links, and labels preserved.}}

\label{tab_metric_audit}
\centering
\small
\setlength{\tabcolsep}{3pt}
\renewcommand{\arraystretch}{1.0}
\begin{adjustbox}{max width=\linewidth}
\begin{tabular*}{\linewidth}{@{\extracolsep{\fill}}lcccccc@{}}
\toprule
Degradation & HPQ$\uparrow$ & OTQ$\uparrow$ & TQ$\uparrow$ & meanNQ$\uparrow$ & MQ$\uparrow$ & LQ$\uparrow$ \\
\midrule
None (original)       & 1.000 & 1.000 & 1.000 & 1.000 & 1.000 & 1.000 \\
Mask erosion          & 0.323 & 0.352 & 0.706 & 0.498 & 0.508 & 0.983 \\
Mask dilation         & 0.670 & 0.640 & 0.979 & 0.653 & 0.668 & 0.978 \\
Parent rewiring       & 0.203 & 0.778 & 0.778 & 1.000 & 1.000 & 1.000 \\
Internal node removal & 0.897 & 0.915 & 0.915 & 1.000 & 1.000 & 1.000 \\
Leaf node removal     & 0.477 & 0.785 & 0.785 & 1.000 & 1.000 & 1.000 \\
Random node removal   & 0.491 & 0.652 & 0.652 & 1.000 & 1.000 & 1.000 \\
Label rephrasing      & 0.311 & 0.982 & 1.000 & 0.982 & 1.000 & 0.982 \\
Label rewiring        & 0.312 & 0.842 & 1.000 & 0.842 & 1.000 & 0.842 \\
\bottomrule
\end{tabular*}
\end{adjustbox}
\end{table}

\subsection{Benchmark Evaluation}
\label{sec_benchmark_tree_comparison}

\rev{We establish baseline results for \datasetname by evaluating different approaches to open tree decomposition on the same 1K-image subset used for human validation. We first compare tree construction strategies while keeping the LVLM and SAM~3 settings fixed. Tree-LVLM predicts the hierarchy in a single pass, either without a predefined depth limit or with a maximum of three levels (object, part, and subpart). Flat-LVLM predicts components without parent-child relations and builds a hierarchy by assigning each component to the smallest larger component that contains it. We then examine the effect of LVLM choice within the \datasetname recursive pipeline, keeping all other settings fixed. We also evaluate HIPIE~\cite{wang2023hipie} using text queries constructed from the node labels and parent-child relations in the \datasetname reference.}

\begin{table*}[h]
\centering
\caption{Comparison of open-tree construction strategies and models. Nodes/img counts semantic nodes, excluding the ``others'' node used for recursive tree construction.}
\small
\setlength{\tabcolsep}{4.2pt}
\renewcommand{\arraystretch}{1.08}
\begin{adjustbox}{max width=\textwidth}
\begin{tabular}{lrrrrrrr}
\toprule
Method & Masks/img & Nodes/img & OTQ$\uparrow$ & TQ$\uparrow$ & meanNQ$\uparrow$ & MQ$\uparrow$ & LQ$\uparrow$ \\
\midrule
\datasetname (reference)
& 83.12 & \rev{25.12} & 1.00 & 1.00 & 1.00 & 1.00 & 1.00 \\
\midrule
\multicolumn{8}{l}{\textit{Different construction strategies (same LVLM)}} \\
Tree-LVLM (no depth limit)
& 12.69 & 8.92 & 0.20 & 0.24 & 0.81 & 0.86 & 0.93 \\
Tree-LVLM (max depth 3)
& 15.32 & 10.27 & 0.25 & 0.30 & 0.84 & 0.88 & 0.96 \\
Flat-LVLM + containment
& 18.53 & 10.73 & 0.27 & 0.32 & 0.82 & 0.87 & 0.95 \\
\midrule
\multicolumn{8}{l}{\textit{Different LVLMs (same recursive pipeline)}} \\
GPT-5.6 Luna
& 62.70 & \rev{19.19} & 0.63 & 0.67 & 0.93 & 0.97 & 0.96 \\
Claude Sonnet 5
& 65.71 & 22.19 & 0.61 & 0.66 & 0.91 & 0.96 & 0.95 \\
Qwen3.7 Plus
& 59.24 & 19.24 & 0.63 & 0.67 & 0.94 & 0.97 & 0.96 \\
Kimi K3
& 64.72 & 20.27 & 0.69 & 0.72 & 0.95 & 0.98 & 0.97 \\
\midrule
\multicolumn{8}{l}{\textit{Existing hierarchical segmentation method}} \\
HIPIE~\cite{wang2023hipie}
& 132.10 & 23.16 & 0.07 & 0.10 & 0.63 & 0.83 & 0.77 \\
\bottomrule
\end{tabular}
\end{adjustbox}
\label{tab:benchmark_results}
\end{table*}

\rev{

Table~\ref{tab:benchmark_results} shows that all evaluated tree construction strategies achieve low OTQ scores (0.20--0.27). Although their meanNQ scores are relatively high (0.81--0.84), they generate far fewer masks than the \datasetname reference (12.69--18.53 vs. 83.12 per image), with TQ scores of only 0.24--0.32. These results suggest that directly predicting a tree (Tree-LVLM) or organizing flat predictions into a hierarchy (Flat-LVLM) is less effective than recursive decomposition at identifying fine-grained components across different depths.

In contrast, changing only the LVLM within the same recursive pipeline produces substantially more masks (59.24--65.71) and semantic nodes (19.19--22.19) per image than the tree construction baselines. All four LVLMs achieve comparable OTQ scores (0.61--0.69) with high matched-node quality (meanNQ: 0.91--0.95). Pairwise comparisons further show that agreement with \datasetname is similar to agreement among the alternative LVLMs (Appendix~\ref{app:pairwise_agreement}). This provides additional support for using \datasetname as a fixed silver reference for benchmarking. We also examine why \datasetname contains more annotations than the alternative LVLM outputs. By comparing proposed and accepted components at each depth, we find that \datasetname generates more proposals at every depth, resulting in more semantic nodes despite a lower proportion of accepted proposals (Appendix~\ref{app:proposal_grounding}). 

Finally, we evaluate HIPIE~\cite{wang2023hipie}, an existing hierarchical segmentation method. Despite receiving reference labels and parent context, HIPIE matches only 10.18 reference masks among 132.10 predicted masks per image, resulting in low TQ (0.10) and OTQ (0.07) scores (Appendix~\ref{app:baseline_details}).

}

\section{Discussion and Limitations}
\label{sec_discussion}

\textbf{Reference multiplicity.}
A fundamental challenge in open tree decomposition is the existence of multiple valid solutions. Because hierarchical parsing is an unconstrained task, a single scene can be logically decomposed in several ways, leading to natural variations in node grouping, intermediate depth, and terminal leaf selection. \rev{\datasetname provides a high-quality silver reference constructed under a consistent visible-evidence annotation policy, with its quality assessed through human review. Other valid trees may receive lower OTQ scores when their components or hierarchy differ from this reference. Our future work includes independently generating and validating additional reference trees for evaluation against multiple valid decompositions.}

\textbf{Open-vocabulary evaluation.} 
Open labels make the benchmark more flexible than fixed category datasets, but they also make label evaluation less direct. The label term in OTQ relies on a label-similarity protocol, which may introduce errors or biases for synonyms, rare words, or parent-dependent meanings. Although we strictly mitigate this lexical ambiguity by evaluating text labels in conjunction with their matched geometric masks and topological tree positions, the inherent fuzziness of automated open-text evaluation remains a structural source of measurement uncertainty.

\textbf{Automated annotation errors.}
Fully automated recursive pipelines inevitably introduce structural and geometric noise. Specifically, the generation process can occasionally miss small child nodes, produce noisy masks, or duplicate redundant instances. These issues are especially prevalent in highly localized, deep-tier regions of the hierarchy. Our comprehensive analysis also revealed these limitations present in \datasetname, thus opening new future research directions for robust, self-verifying automated annotation frameworks.

\section{Conclusion}
\label{sec_conclusion}
We introduced \datasetname, a COCO-based dataset and benchmark for \emph{open tree decomposition}. It represents each image as an open-vocabulary tree of visible instances, where each node is grounded by an instance mask and a local semantic label. This moves segmentation beyond flat region prediction and fixed object-part templates toward image-specific structural understanding.
We constructed \datasetname with a fully automated recursive annotation pipeline that uses LVLMs for parent-conditioned child proposal and the vision foundation model (SAM~3) for mask grounding. The dataset contains over \numImages images, \numMasks instance nodes (85.7 masks per image), and more than 3.5K open-vocabulary labels. Human validation supports the quality of its masks, labels, and hierarchical structures. We also proposed Open Tree Quality (OTQ), a PQ-style metric that jointly evaluates node recovery, mask quality, label quality, and visual parent consistency. Together, \datasetname and OTQ establish a benchmark for models that must recover not only visible regions but also their compositional organization. We believe this setting opens a path toward segmentation systems that better support fine-grained reasoning, interaction, and embodied visual understanding.

\textbf{Broader impacts.}
As with any automatically constructed visual dataset, \datasetname should not be treated as a complete or error-free description of visual structure. Models trained on it may inherit annotation errors, dataset biases, or failures on small and ambiguous regions.

\clearpage
\begin{ack}
This work was supported by the Korea Health Technology R\&D Project through the Korea Health Industry Development Institute (KHIDI), funded by the Ministry of Health and Welfare, Republic of Korea (RS-2026-25506103);
in part by the National Research Foundation of Korea (NRF) under grants RS-2025-02217071 and RS-2025-00520248;
and in part by the Institute of Information \& Communications Technology Planning \& Evaluation (IITP), funded by the Korean government (MSIT), under grants RS-2025-02305436 (Development of Digital Innovative Element Technologies for Rapid Prediction of Potential Complex Disasters and Continuous Disaster Prevention) and IITP-2026-RS-2026-25545512 (Leading Generative AI Human Resources Development).

The authors declare no competing interests.
\end{ack}

\bibliographystyle{plainnat}
\bibliography{reference}

\clearpage 
\appendix
\section{Prompt Details}
\label{app:prompts}

This appendix summarizes the prompts used by the LVLM planner in our construction pipeline.
The prompts are designed to produce visible, recognizable, separable, and structurally meaningful child labels while avoiding hidden or merely expected parts.

\subsection{System Prompt}
\label{app:system_prompt}

\paragraph{Objective.} The system prompt asks the planner to produce a rich hierarchy of visible structure.
It encourages decomposition whenever clear and visually supported subparts exist.
It also prefers continuing decomposition over stopping early.

\paragraph{Evidence policy.}
\begin{itemize}[leftmargin=1.25em, itemsep=0.2em]
    \item Use only visible evidence from the image or masked crop.
    \item Do not infer hidden, occluded, or merely expected parts.
    \item Propose a candidate only when it is visually supported and reasonably separable.
    \item Prefer decompositions that reveal more meaningful visible structure.
    \item If an intermediate part is unclear but a finer part is clearly visible, propose the finer part directly.
\end{itemize}

\paragraph{Hierarchy policy.}
\begin{itemize}[leftmargin=1.25em, itemsep=0.2em]
    \item Continue decomposition whenever another meaningful visible level can be exposed.
    \item Require each child to be a meaningful subpart of its parent.
    \item Prefer structural parts, functional parts, attached components, and clearly bounded visible regions.
    \item Prefer larger structural parts when they are clear, while still allowing smaller visible parts.
    \item Prefer intermediate levels when they are visible and separable, but do not force them when they are unclear.
    \item Avoid outputting a parent like concept and its likely internal subpart in the same step.
    \item Stop only when no further clear, recognizable, visually separable, and structurally meaningful subpart can be proposed.
\end{itemize}

\paragraph{Rejection policy.}
\begin{itemize}[leftmargin=1.25em, itemsep=0.2em]
    \item Do not propose materials, textures, colors, patterns, lighting effects, reflections, shadows, or abstract concepts.
    \item Do not propose labels that are too vague or that merely restate the parent or an ancestor.
    \item Do not invent nonvisible parts.
    \item Do not output duplicates, near synonyms, or overlapping alternatives for the same region.
    \item Do not propose boundary only regions such as rims, borders, outlines, edges, or outer rings unless they are distinct physical components.
    \item Do not split an object into an outer band and inner area when the split follows only silhouette, perimeter, or graphic layout.
\end{itemize}

\paragraph{Label policy.}
\begin{itemize}[leftmargin=1.25em, itemsep=0.2em]
    \item Use singular concrete nouns.
    \item Prefer common everyday words.
    \item Prefer one word and use two words only for common compound nouns.
    \item Avoid adjectives, attributes, colors, materials, positions, lighting terms, and abstract words.
    \item Avoid punctuation, quotes, and duplicate labels.
    \item If labels overlap strongly, keep one label and prefer the more common term unless the finer term reveals a meaningful decomposition level.
    \item Avoid technical, scientific, anatomical, or highly specialized terms.
\end{itemize}

\subsection{Initial Root Discovery Prompt}
\label{app:root_prompt}

The initial discovery prompt proposes root level anchors that maximize visible coverage and support later decomposition.
It is intentionally biased toward larger visible units because internal parts can be discovered in later parent conditioned steps.

\begin{itemize}[leftmargin=1.25em, itemsep=0.2em]
    \item Propose root level units only.
    \item Prefer whole objects and major scene regions that can act as strong anchors.
    \item Include dominant scene regions such as sky, ground, water, road, mountain, wall, floor, and ceiling when clearly visible.
    \item Include distinct whole objects such as person, car, building, animal, furniture, plant, container, tool, and appliance.
    \item Maximize coverage of major visible structure.
    \item Prefer candidates that enable rich later decomposition.
    \item Do not propose internal parts when a larger visible object can serve as the root anchor.
    \item If one candidate is likely a part of another visible candidate, propose only the larger candidate at this step.
    \item Avoid tiny fragments, textures, materials, coverings, duplicates, synonyms, positional variants, and overly abstract scene labels.
    \item Propose enough roots for strong coverage and later decomposition, typically between four and twenty four candidates.
\end{itemize}

\subsection{Local Decomposition Prompt}
\label{app_local_prompt}

The local decomposition prompt expands multiple accepted parent nodes in a batch.
For each parent, the LVLM receives the masked crop and current path, then proposes child labels using only the masked crop as visual evidence.

\begin{itemize}[leftmargin=1.25em, itemsep=0.2em]
    \item Process each parent independently.
    \item Use the path and label only for filtering and context, not for inventing unseen parts.
    \item Do not stop if clear subparts exist.
    \item Propose as many meaningful visible subparts as can be supported, up to the configured maximum number of children.
    \item Prefer structural parts, functional parts, attached components, and clearly bounded visible regions.
    \item Prefer intermediate levels when they are clearly supported and separable.
    \item If an intermediate level is unclear but a finer part is visible and commonly recognized, propose the finer part directly.
    \item Keep children at a similar structural level when possible.
    \item Allow attached accessories, appendages, and externally visible components as direct children.
    \item Avoid materials, textures, colors, patterns, coverings, and surface layers.
    \item Return an empty child list when no clear, recognizable, visually separable, and meaningful subpart can be proposed.
\end{itemize}

\subsection{Output Contract}
\label{app:output_contract}

The prompt requires structured tool output so that proposals can be parsed and audited automatically. For initial discovery, the output is a list of root text prompts. For local decomposition, the output includes every parent identifier exactly once and assigns each parent either a list of child prompts or an empty list. The planner is instructed to return only the tool call and no free form explanation.

\section{SAM 3 Confidence Threshold Sensitivity}
\label{app:confidence_threshold}

We evaluate the sensitivity of the SAM~3 confidence threshold on the 1K-image human-validation subset using thresholds of $0.4$, $0.5$, $0.6$, and $0.7$.
Table~\ref{tab:confidence_threshold} reports the average number of retained semantic nodes per image and mask average recall (AR) against four external reference datasets.

\begin{table*}[h]
\centering
\caption{
SAM~3 confidence threshold sensitivity analysis.
We report the average number of retained semantic nodes per image and mask average recall (AR) across four external references, stratified by mask size.
}
\small
\setlength{\tabcolsep}{4pt}
\renewcommand{\arraystretch}{1.08}
\begin{adjustbox}{max width=\textwidth}
\begin{tabular}{clcccccc}
\toprule
Threshold & Reference & Nodes/img & AR$\uparrow$ & AR$_{\mathrm{XS}}\uparrow$ & AR$_{S^\star}\uparrow$ & AR$_M\uparrow$ & AR$_L\uparrow$ \\
\midrule
\multirow{5}{*}{0.4}
& COCO        & \multirow{5}{*}{25.08} & 0.59 & 0.17 & 0.48 & 0.66 & 0.75 \\
& COCO-ReM    &                        & 0.68 & 0.21 & 0.55 & 0.81 & 0.89 \\
& COCONut     &                        & 0.62 & 0.20 & 0.47 & 0.70 & 0.85 \\
& LVIS        &                        & 0.48 & 0.09 & 0.42 & 0.68 & 0.85 \\
& \textbf{Average} &                    & \textbf{0.59} & \textbf{0.17} & \textbf{0.48} & \textbf{0.71} & \textbf{0.83} \\
\midrule

\multirow{5}{*}{0.5}
& COCO        & \multirow{5}{*}{25.12} & 0.58 & 0.13 & 0.47 & 0.65 & 0.74 \\
& COCO-ReM    &                        & 0.65 & 0.16 & 0.51 & 0.80 & 0.88 \\
& COCONut     &                        & 0.59 & 0.16 & 0.43 & 0.69 & 0.84 \\
& LVIS        &                        & 0.46 & 0.08 & 0.39 & 0.67 & 0.83 \\
& \textbf{Average} &                    & \textbf{0.57} & \textbf{0.13} & \textbf{0.45} & \textbf{0.70} & \textbf{0.82} \\
\midrule

\multirow{5}{*}{0.6}
& COCO        & \multirow{5}{*}{24.67} & 0.57 & 0.11 & 0.45 & 0.64 & 0.73 \\
& COCO-ReM    &                        & 0.63 & 0.13 & 0.48 & 0.79 & 0.87 \\
& COCONut     &                        & 0.57 & 0.13 & 0.40 & 0.68 & 0.83 \\
& LVIS        &                        & 0.44 & 0.07 & 0.38 & 0.64 & 0.81 \\
& \textbf{Average} &                    & \textbf{0.55} & \textbf{0.11} & \textbf{0.43} & \textbf{0.69} & \textbf{0.81} \\
\midrule

\multirow{5}{*}{0.7}
& COCO        & \multirow{5}{*}{24.06} & 0.55 & 0.09 & 0.42 & 0.62 & 0.72 \\
& COCO-ReM    &                        & 0.60 & 0.10 & 0.43 & 0.77 & 0.86 \\
& COCONut     &                        & 0.55 & 0.10 & 0.35 & 0.67 & 0.82 \\
& LVIS        &                        & 0.43 & 0.07 & 0.36 & 0.65 & 0.79 \\
& \textbf{Average} &                    & \textbf{0.53} & \textbf{0.09} & \textbf{0.39} & \textbf{0.68} & \textbf{0.80} \\
\bottomrule
\end{tabular}
\end{adjustbox}
\label{tab:confidence_threshold}
\end{table*}

The threshold of $0.5$ retains the highest average number of semantic nodes per image (25.12), although the difference from $0.4$ (25.08) is small. Mean AR across the four references decreases as the threshold increases, particularly for small masks: AR for XS and S$^\star$ masks drops from $0.17$ to $0.09$ and from $0.48$ to $0.39$, respectively.

\section{Small Mask Overlap Analysis}
\label{app:small_mask_overlap}

We analyze the overlap between \datasetname and four external reference datasets (COCO-17, COCONut, COCO-ReM, and LVIS) for Small ($S^\star$) and Extra Small (XS) masks. Table~\ref{tab:small_mask_overlap} reports the matching rates in both directions and the mean IoU of matched masks. External matched (\%) measures the coverage of external reference masks by \datasetname, while \datasetname matched (\%) measures the coverage of \datasetname masks by each external reference.

\begin{table*}[h]
\centering
\caption{Bidirectional small-mask overlap between \datasetname and external reference datasets. Matching rates are reported separately for external reference masks and \datasetname masks. Matched IoU denotes the mean IoU of matched mask pairs.}
\small
\setlength{\tabcolsep}{4pt}
\renewcommand{\arraystretch}{1.08}
\begin{adjustbox}{max width=\textwidth}
\begin{tabular}{llrrrrrr}
\toprule
\multirow{2}{*}{Reference} &
\multirow{2}{*}{Size} &
\multicolumn{2}{c}{External reference} &
\multicolumn{2}{c}{\datasetname} &
\multirow{2}{*}{Matched masks} &
\multirow{2}{*}{Matched IoU} \\
\cmidrule(lr){3-4}
\cmidrule(lr){5-6}
& & Masks & Matched (\%) & Masks & Matched (\%) & & \\
\midrule
\multirow{2}{*}{COCO-17}
& $S^\star$ & 1,767 & 74.2 & 71,209 & 1.89 & 1,311 & 0.788 \\
& XS        & 445   & 28.3 & 29,002 & 0.44 & 126   & 0.706 \\
\midrule

\multirow{2}{*}{COCONut}
& $S^\star$ & 3,203 & 69.7 & 74,431 & 3.06 & 2,232 & 0.781 \\
& XS        & 285   & 34.7 & 26,351 & 0.38 & 99    & 0.706 \\
\midrule

\multirow{2}{*}{COCO-ReM}
& $S^\star$ & 2,941 & 64.5 & 76,247 & 2.53 & 1,896 & 0.861 \\
& XS        & 701   & 25.1 & 32,166 & 0.54 & 176   & 0.800 \\
\midrule

\multirow{2}{*}{LVIS}
& $S^\star$ & 3,954 & 56.2 & 77,813 & 2.86 & 2,222 & 0.809 \\
& XS        & 1,861 & 16.4 & 52,089 & 0.59 & 305   & 0.726 \\
\bottomrule
\end{tabular}
\end{adjustbox}
\label{tab:small_mask_overlap}
\end{table*}

For XS masks, only 16.4--34.7\% of external reference masks are matched, while the matched pairs achieve mean IoUs of 0.706--0.800.
Conversely, each external reference covers only 0.38--0.59\% of \datasetname's XS masks. These results indicate substantial differences in annotation coverage, as \datasetname includes fine-grained parts and subparts that are generally absent from the external references. Thus, lower recall for small masks is influenced by limited annotation overlap and does not necessarily indicate poor mask quality. Human validation further supports the quality of deeper annotations, with Q5 (boundary quality) Good-or-Minor rates of 96.06\% at depth 3 and 95.43\% at depth $\geq4$.

\section{Human Validation Website}
\label{app:review_web}

\begin{figure*}[t]
\centering
\includegraphics[width=\textwidth]{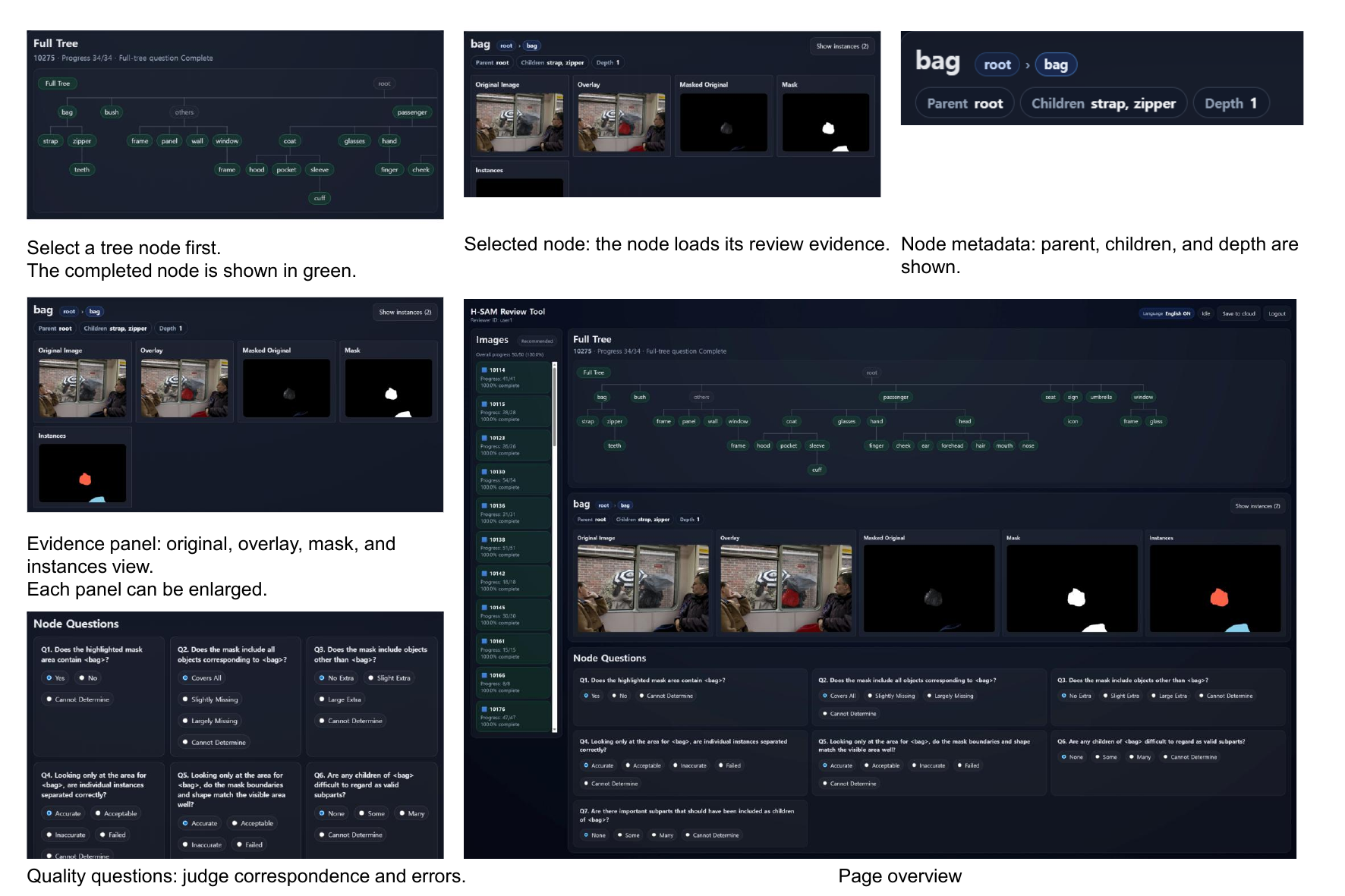}
\caption{Human validation website showing the image, annotation hierarchy, highlighted masks, and evaluation questions.}
\label{fig_review_web}
\end{figure*}

Figure~\ref{fig_review_web} shows the website used by human evaluators to assess \datasetname annotations.
For each semantic node, the interface displayed the original image, parent context, highlighted masks, local label, and current tree path.

Evaluators answered Q1--Q5 for every evaluated node.
Non-leaf nodes were additionally assessed using Q6--Q7 for child validity and completeness, while leaf nodes were assessed using Q8 for stopping appropriateness.
After completing the node-level questions for an image, evaluators answered the full-tree questions T1--T2.
Table~\ref{tab:human_validation_questions} lists the questions and response options, where \texttt{<Label>} represents the local label of the evaluated node.

\begin{table*}[t]
\centering
\caption{Questions and response options used in human validation.}
\label{tab:human_validation_questions}
\small
\setlength{\tabcolsep}{3.5pt}
\renewcommand{\arraystretch}{1.12}
\begin{adjustbox}{max width=\textwidth}
\begin{tabularx}{\textwidth}{L{0.75cm} L{1.6cm} L{2.7cm} Y L{4.0cm}}
\toprule
ID & Scope & Axis & Question shown to evaluators & Response options \\
\midrule
Q1 & All nodes & Label-mask match & Does the highlighted mask area contain \texttt{<Label>}? & Yes; No; Cannot determine \\
Q2 & All nodes & Target coverage & Does the mask include all objects corresponding to \texttt{<Label>}? & Covers all; Slightly missing; Largely missing; Cannot determine \\
Q3 & All nodes & Extra region & Does the mask include objects other than \texttt{<Label>}? & No extra; Slight extra; Large extra; Cannot determine \\
Q4 & All nodes & Instance separation & Looking only at the area for \texttt{<Label>}, are individual instances separated correctly? & Accurate; Acceptable; Inaccurate; Failed; Cannot determine \\
Q5 & All nodes & Boundary quality & Looking only at the area for \texttt{<Label>}, do the mask boundaries and shape match the visible area well? & Accurate; Acceptable; Inaccurate; Failed; Cannot determine \\
\midrule
Q6 & Non-leaf only & Child validity & Are any children of \texttt{<Label>} difficult to regard as valid subparts? & None; Some; Many; Cannot determine \\
Q7 & Non-leaf only & Missing children & Are there important subparts that should have been included as children of \texttt{<Label>}? & None; Some; Many; Cannot determine \\
\midrule
Q8 & Leaf only & Leaf stopping & Is it appropriate not to subdivide \texttt{<Label>} any further here? & Yes; No; Cannot determine \\
\midrule
T1 & Full tree & Tree consistency & Does the full tree appropriately decompose the image? & Correct; Mostly correct; Partly correct; Incorrect; Cannot determine \\
T2 & Full tree & Remaining area & Are there meaningful elements in the remaining area? & None; Some; Many; Cannot determine \\
\bottomrule
\end{tabularx}
\end{adjustbox}
\end{table*}
\section{Boundary Quality by Semantic Depth}
\label{app:depth_validation}

We further analyze boundary quality across semantic depths using the human evaluation results for Q5 (Boundary quality). Table~\ref{tab:boundary_depth_validation} shows the response distributions at each depth.

\begin{table*}[h]
\centering
\caption{Human evaluation results for Q5 by semantic depth. Values are percentages with standard deviations.}
\small
\setlength{\tabcolsep}{4.2pt}
\renewcommand{\arraystretch}{1.08}
\begin{adjustbox}{max width=\textwidth}
\begin{tabular}{lccccc}
\toprule
Depth & Good (\%) & Minor (\%) & Moderate (\%) & Major (\%) & Not clear (\%) \\
\midrule
1 & $82.93 \pm 10.10$ & $14.87 \pm 9.20$  & $2.03 \pm 1.75$ & $0.10 \pm 0.16$ & $0.07 \pm 0.17$ \\
2 & $77.84 \pm 11.10$ & $16.97 \pm 9.94$  & $4.38 \pm 2.46$ & $0.62 \pm 1.12$ & $0.19 \pm 0.30$ \\
3 & $78.31 \pm 12.62$ & $17.75 \pm 11.58$ & $3.40 \pm 2.72$ & $0.44 \pm 0.80$ & $0.11 \pm 0.23$ \\
$\geq 4$ & $75.37 \pm 16.04$ & $20.06 \pm 14.02$ & $3.74 \pm 2.75$ & $0.54 \pm 1.02$ & $0.30 \pm 0.78$ \\
\bottomrule
\end{tabular}
\end{adjustbox}
\label{tab:boundary_depth_validation}
\end{table*}

Boundary quality remains broadly stable across semantic depths. Good ratings decrease from $82.93\%$ at depth 1 to $75.37\%$ at depth $\geq 4$, with most of the difference shifting to Minor ratings. The combined Good-or-Minor rate remains above $94.8\%$ at every depth, reaching $96.06\%$ at depth 3 and $95.43\%$ at depth $\geq 4$. Major ratings remain at or below $0.62\%$ across all depths. These results indicate that deeper decomposition maintains comparable boundary quality despite the increased prevalence of small masks.

\section{Node Matching Threshold Analysis}
\label{app:iou_analysis}

We further examine the node matching threshold of $\tau_{\mathrm{node}}=0.5$ using the human-annotated or human-reviewed reference datasets in Table~\ref{tab:mask_compatibility}. For each reference mask, we compute its best-match IoU against \datasetname masks. Table~\ref{tab:best_match_iou} reports the mean and percentiles of these IoUs, along with Match@$0.5$, the percentage of reference masks whose best-match IoU is at least $0.5$.

\begin{table*}[h]
\centering
\caption{Best-match IoU distributions between \datasetname and human-annotated reference datasets. Match@$0.5$ denotes the percentage of reference masks with a best-match IoU of at least $0.5$.}
\small
\setlength{\tabcolsep}{4.2pt}
\renewcommand{\arraystretch}{1.08}
\begin{adjustbox}{max width=\textwidth}
\begin{tabular}{lccccccc}
\toprule
Reference & Mean & P10 & P25 & Median & P75 & P90 & Match@$0.5$ \\
\midrule
COCO-17  & 0.69 & 0.03 & 0.62 & 0.81 & 0.90 & 0.94 & 80.2\% \\
COCONut  & 0.69 & 0.02 & 0.63 & 0.83 & 0.91 & 0.95 & 79.3\% \\
COCO-ReM & 0.69 & 0.00 & 0.47 & 0.90 & 0.96 & 0.98 & 74.4\% \\
LVIS     & 0.54 & 0.00 & 0.03 & 0.73 & 0.90 & 0.95 & 59.3\% \\
\bottomrule
\end{tabular}
\end{adjustbox}
\label{tab:best_match_iou}
\end{table*}

Across the four references, median best-match IoUs range from $0.73$ to $0.90$, with $59.3$--$80.2\%$ of reference masks achieving IoU $\geq 0.5$. These results support $0.5$ as a practical threshold for identifying spatially corresponding masks. Higher thresholds impose stricter matching criteria but reduce match coverage.
\section{Dataset Distribution}
\label{app:data_distribution}

\begin{figure*}[h]
\centering
\includegraphics[width=\textwidth]{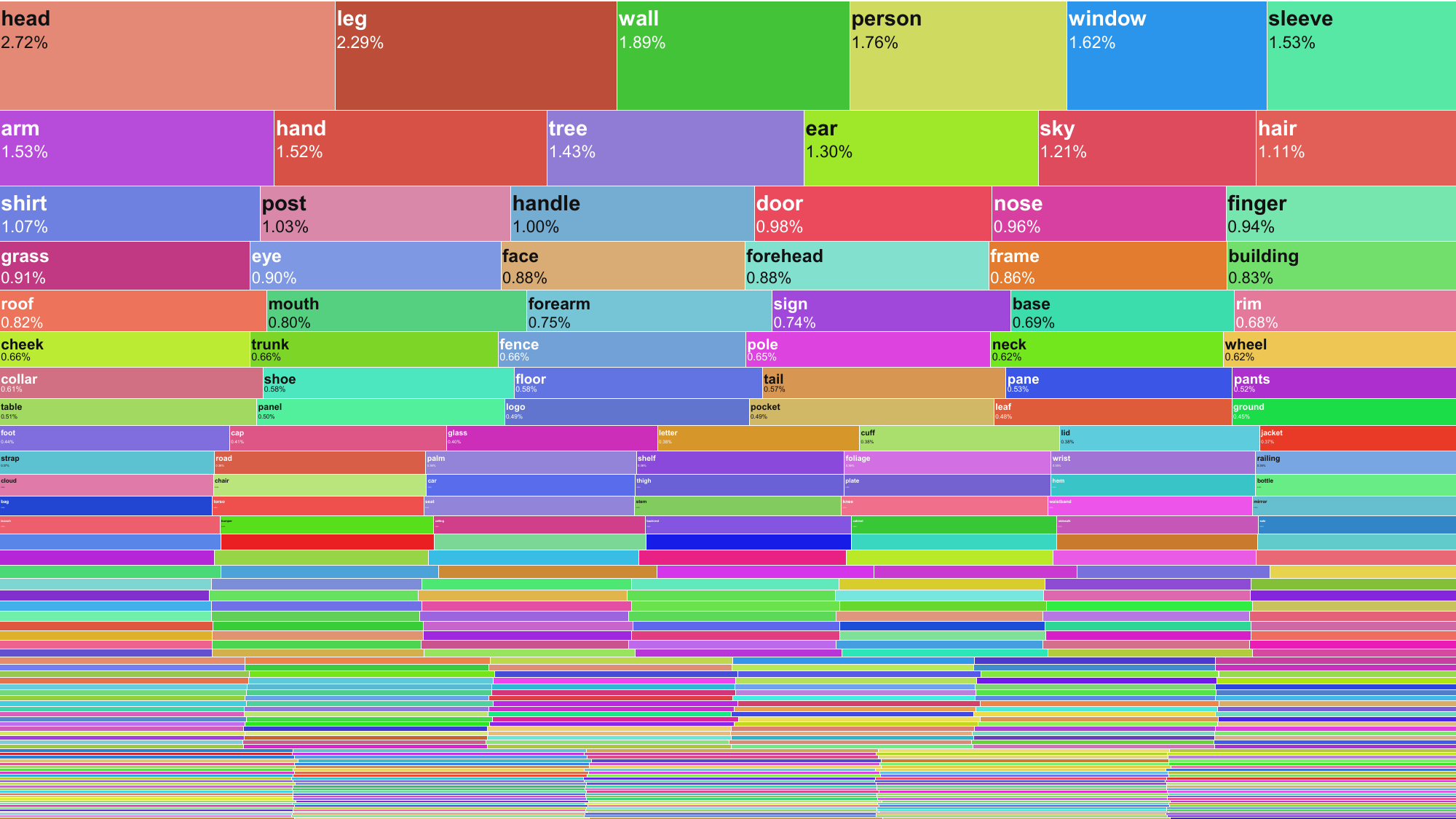}
\vspace{-0.2cm}
\caption{Label distribution treemap for \datasetname.}
\label{fig:label_distribution}
\end{figure*}

Figure~\ref{fig:label_distribution} shows the label distribution of \datasetname.
Only annotations that passed the integrity check are included, excluding nodes whose terminal label is \texttt{others}.
Each rectangle represents a label, with its area proportional to the frequency of that label.

\begin{figure*}[h]
\centering
\includegraphics[width=\textwidth]{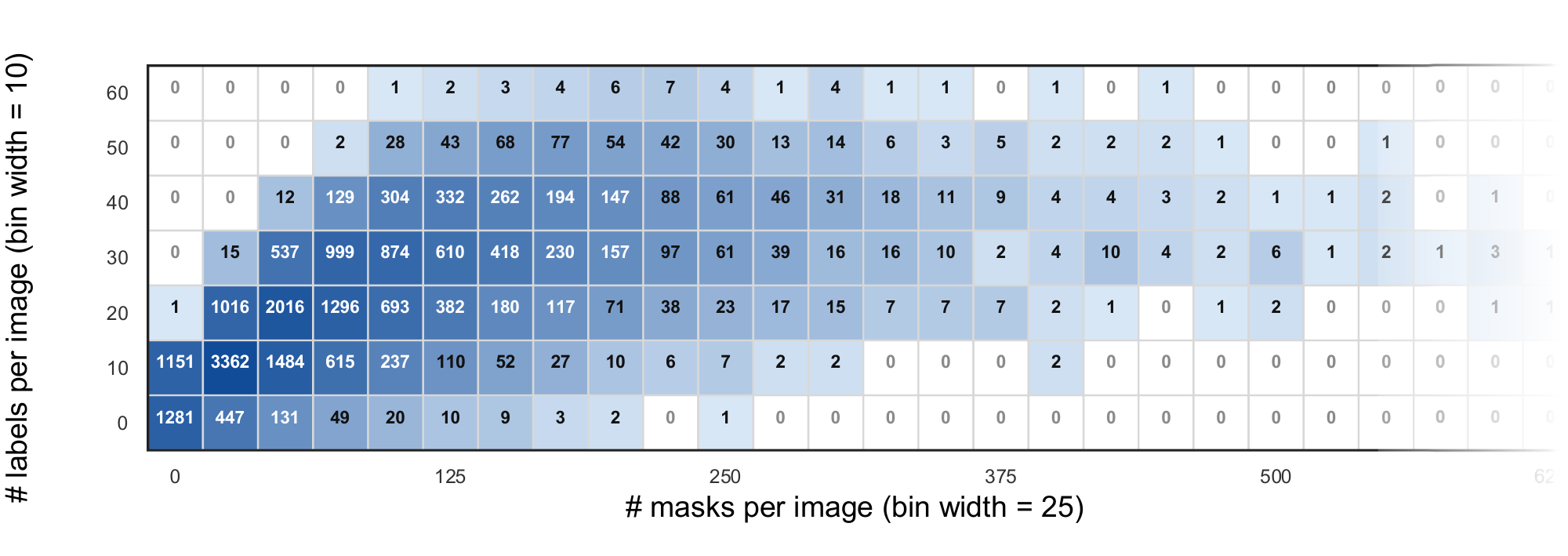}
\vspace{-0.2cm}
\caption{Joint distribution of masks and unique labels per image in \datasetname. Mask counts are grouped into bins of 25, and unique label counts into bins of 10.}
\label{fig:mask_vs_label_distribution}
\end{figure*}

Figure~\ref{fig:mask_vs_label_distribution} shows the joint distribution of masks and unique labels per image.
Each cell represents the number of images within a given range of mask and label counts.

\begin{figure}[h]
\centering
\includegraphics[width=\linewidth]{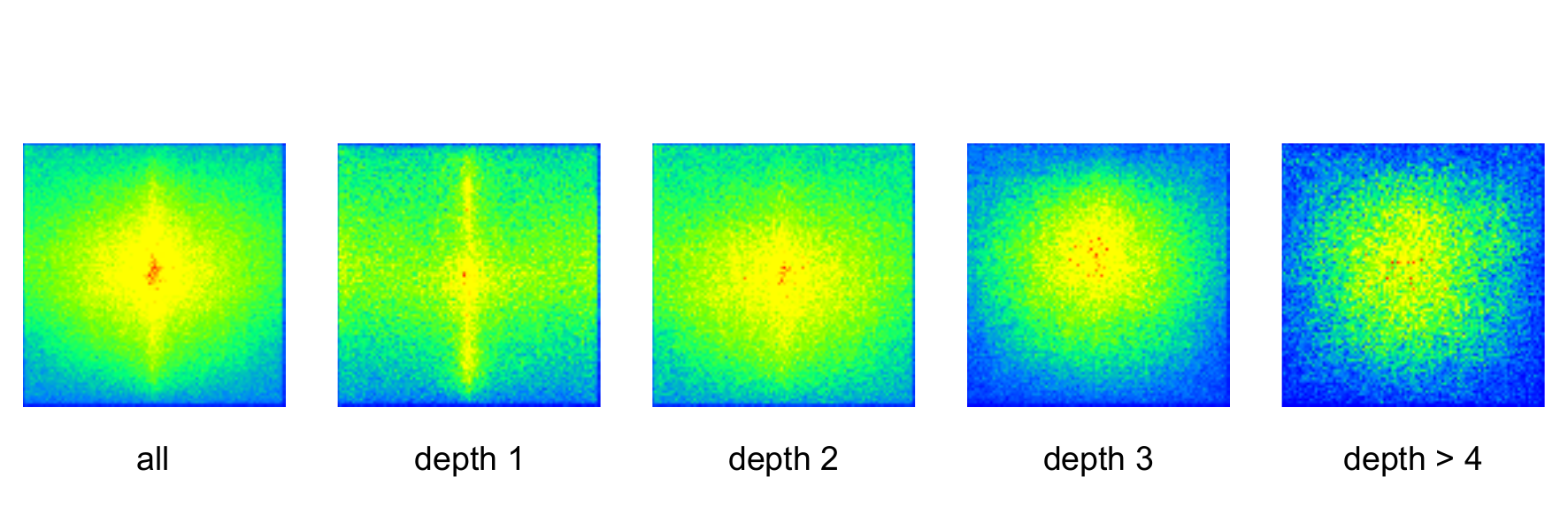}
\caption{Spatial distribution of annotation centers in the human validation subset.}
\label{fig_center_distribution}
\end{figure}

Figure~\ref{fig_center_distribution} shows the spatial distribution of annotation centers in the human-validation subset, illustrating where annotated regions are concentrated across the image plane.

\section{Additional OTQ Evaluation Results}
\label{app:otq_additional_results}

We provide additional controlled degradation results for \datasetname annotations and recursively generated SAM 3 outputs, along with an analysis of label similarity methods.
The degradation experiments use the same 1K-image subset as Section~\ref{sec_metric_audit}, with BERT-based label similarity.

\subsection{\datasetname Annotation Degradations}
\label{app:gt_degradation}

Table~\ref{tab:appendix_gt_degradation_full} extends Table~\ref{tab_metric_audit} to keep ratios of 75\%, 50\%, 30\%, and 15\%.
Lower keep ratios indicate stronger degradations. OTQ consistently decreases as degradation severity increases, with its components reflecting the corresponding geometric, structural, and semantic changes.

\begin{table*}[h]
\caption{Evaluation of OTQ under controlled degradations of \datasetname annotations at different keep ratios.}
\label{tab:appendix_gt_degradation_full}
\centering
\small
\setlength{\tabcolsep}{3.2pt}
\renewcommand{\arraystretch}{1.08}
\begin{adjustbox}{max width=\textwidth}
\begin{tabular}{llcccccc}
\toprule
Degradation & Keep ratio (\%) & HPQ$\uparrow$ & OTQ$\uparrow$ & TQ$\uparrow$ & meanNQ$\uparrow$ & MQ$\uparrow$ & LQ$\uparrow$ \\
\midrule
None (original) & -- & 1.000 & 1.000 & 1.000 & 1.000 & 1.000 & 1.000 \\
\midrule

\multirow{4}{*}{Mask erosion}
& 75 & 0.750 & 0.731 & 0.990 & 0.738 & 0.750 & 0.984 \\
& 50 & 0.323 & 0.352 & 0.706 & 0.498 & 0.508 & 0.983 \\
& 30 & 0.000 & 0.004 & 0.009 & 0.195 & 0.233 & 0.343 \\
& 15 & 0.000 & 0.001 & 0.002 & 0.058 & 0.075 & 0.100 \\
\midrule

\multirow{4}{*}{Mask dilation}
& 75 & 0.801 & 0.780 & 0.990 & 0.788 & 0.800 & 0.984 \\
& 50 & 0.670 & 0.640 & 0.979 & 0.653 & 0.668 & 0.978 \\
& 30 & 0.594 & 0.556 & 0.963 & 0.578 & 0.592 & 0.977 \\
& 15 & 0.547 & 0.503 & 0.943 & 0.533 & 0.547 & 0.977 \\
\midrule

\multirow{4}{*}{Parent rewiring}
& 75 & 0.463 & 0.872 & 0.872 & 1.000 & 1.000 & 1.000 \\
& 50 & 0.203 & 0.778 & 0.778 & 1.000 & 1.000 & 1.000 \\
& 30 & 0.100 & 0.729 & 0.729 & 1.000 & 1.000 & 1.000 \\
& 15 & 0.045 & 0.692 & 0.692 & 1.000 & 1.000 & 1.000 \\
\midrule

\multirow{4}{*}{Internal node removal}
& 75 & 0.951 & 0.964 & 0.965 & 1.000 & 1.000 & 1.000 \\
& 50 & 0.897 & 0.915 & 0.915 & 1.000 & 1.000 & 1.000 \\
& 30 & 0.863 & 0.873 & 0.874 & 1.000 & 1.000 & 1.000 \\
& 15 & 0.842 & 0.842 & 0.842 & 1.000 & 1.000 & 1.000 \\
\midrule

\multirow{4}{*}{Leaf node removal}
& 75 & 0.747 & 0.899 & 0.899 & 1.000 & 1.000 & 1.000 \\
& 50 & 0.477 & 0.785 & 0.785 & 1.000 & 1.000 & 1.000 \\
& 30 & 0.268 & 0.673 & 0.673 & 1.000 & 1.000 & 1.000 \\
& 15 & 0.128 & 0.571 & 0.571 & 1.000 & 1.000 & 1.000 \\
\midrule

\multirow{4}{*}{Random node removal}
& 75 & 0.712 & 0.853 & 0.854 & 1.000 & 1.000 & 1.000 \\
& 50 & 0.491 & 0.652 & 0.652 & 1.000 & 1.000 & 1.000 \\
& 30 & 0.303 & 0.451 & 0.451 & 1.000 & 1.000 & 1.000 \\
& 15 & 0.146 & 0.263 & 0.263 & 0.996 & 0.996 & 0.996 \\
\midrule

\multirow{4}{*}{Label rephrasing}
& 75 & 0.550 & 0.991 & 1.000 & 0.991 & 1.000 & 0.991 \\
& 50 & 0.311 & 0.982 & 1.000 & 0.982 & 1.000 & 0.982 \\
& 30 & 0.188 & 0.975 & 1.000 & 0.975 & 1.000 & 0.975 \\
& 15 & 0.127 & 0.969 & 1.000 & 0.969 & 1.000 & 0.969 \\
\midrule

\multirow{4}{*}{Label rewiring}
& 75 & 0.551 & 0.920 & 1.000 & 0.920 & 1.000 & 0.920 \\
& 50 & 0.312 & 0.842 & 1.000 & 0.842 & 1.000 & 0.842 \\
& 30 & 0.189 & 0.776 & 1.000 & 0.776 & 1.000 & 0.776 \\
& 15 & 0.128 & 0.730 & 1.000 & 0.730 & 1.000 & 0.730 \\
\bottomrule
\end{tabular}
\end{adjustbox}
\end{table*}

\subsection{Recursive SAM 3 Output Degradations}
\label{app:recursive_corruption_results}

We additionally evaluate OTQ on open trees generated by recursively applying SAM 3 within cropped semantic regions.
At each step, the bounding box of a semantic region is cropped, and SAM 3 is applied within the crop to obtain finer component masks. This process is repeated to construct multilevel trees, which are evaluated against the original \datasetname annotations. We then apply the same controlled degradations to these generated trees. Table~\ref{tab:appendix_recursive_degradation_full} shows that OTQ consistently decreases with increasing degradation severity.

\begin{table*}[t]
\caption{Evaluation of OTQ under controlled degradations of recursive SAM 3-based output trees. All outputs are evaluated against the original \datasetname annotations.}
\label{tab:appendix_recursive_degradation_full}
\centering
\small
\setlength{\tabcolsep}{3.2pt}
\renewcommand{\arraystretch}{1.08}
\begin{adjustbox}{max width=\textwidth}
\begin{tabular}{llcccccc}
\toprule
Degradation & Keep ratio (\%) & HPQ$\uparrow$ & OTQ$\uparrow$ & TQ$\uparrow$ & meanNQ$\uparrow$ & MQ$\uparrow$ & LQ$\uparrow$ \\
\midrule
None (unmodified) & -- & 0.349 & 0.587 & 0.655 & 0.895 & 0.901 & 0.994 \\
\midrule

\multirow{4}{*}{Mask erosion}
& 75 & 0.247 & 0.469 & 0.644 & 0.727 & 0.740 & 0.982 \\
& 50 & 0.075 & 0.208 & 0.398 & 0.521 & 0.537 & 0.969 \\
& 30 & 0.000 & 0.005 & 0.009 & 0.185 & 0.213 & 0.330 \\
& 15 & 0.000 & 0.001 & 0.002 & 0.036 & 0.046 & 0.062 \\
\midrule

\multirow{4}{*}{Mask dilation}
& 75 & 0.279 & 0.481 & 0.640 & 0.752 & 0.763 & 0.985 \\
& 50 & 0.219 & 0.397 & 0.620 & 0.640 & 0.653 & 0.982 \\
& 30 & 0.181 & 0.336 & 0.585 & 0.575 & 0.587 & 0.980 \\
& 15 & 0.154 & 0.281 & 0.523 & 0.537 & 0.549 & 0.979 \\
\midrule

\multirow{4}{*}{Parent rewiring}
& 75 & 0.157 & 0.528 & 0.589 & 0.895 & 0.901 & 0.994 \\
& 50 & 0.071 & 0.469 & 0.522 & 0.895 & 0.901 & 0.994 \\
& 30 & 0.031 & 0.431 & 0.481 & 0.895 & 0.901 & 0.994 \\
& 15 & 0.012 & 0.405 & 0.452 & 0.895 & 0.901 & 0.994 \\
\midrule

\multirow{4}{*}{Internal node removal}
& 75 & 0.343 & 0.560 & 0.628 & 0.891 & 0.898 & 0.992 \\
& 50 & 0.339 & 0.525 & 0.592 & 0.885 & 0.893 & 0.991 \\
& 30 & 0.333 & 0.497 & 0.561 & 0.884 & 0.893 & 0.990 \\
& 15 & 0.331 & 0.475 & 0.540 & 0.879 & 0.888 & 0.989 \\
\midrule

\multirow{4}{*}{Leaf node removal}
& 75 & 0.250 & 0.523 & 0.580 & 0.901 & 0.907 & 0.994 \\
& 50 & 0.165 & 0.444 & 0.490 & 0.905 & 0.910 & 0.993 \\
& 30 & 0.094 & 0.363 & 0.398 & 0.907 & 0.913 & 0.989 \\
& 15 & 0.049 & 0.309 & 0.339 & 0.903 & 0.909 & 0.984 \\
\midrule

\multirow{4}{*}{Random node removal}
& 75 & 0.267 & 0.484 & 0.541 & 0.893 & 0.900 & 0.993 \\
& 50 & 0.191 & 0.356 & 0.403 & 0.881 & 0.891 & 0.987 \\
& 30 & 0.111 & 0.246 & 0.272 & 0.900 & 0.908 & 0.989 \\
& 15 & 0.059 & 0.130 & 0.150 & 0.842 & 0.854 & 0.962 \\
\midrule

\multirow{4}{*}{Label rephrasing}
& 75 & 0.205 & 0.582 & 0.655 & 0.887 & 0.901 & 0.985 \\
& 50 & 0.115 & 0.577 & 0.655 & 0.880 & 0.901 & 0.977 \\
& 30 & 0.061 & 0.573 & 0.655 & 0.874 & 0.901 & 0.970 \\
& 15 & 0.028 & 0.569 & 0.655 & 0.869 & 0.901 & 0.964 \\
\midrule

\multirow{4}{*}{Label rewiring}
& 75 & 0.206 & 0.539 & 0.655 & 0.823 & 0.901 & 0.913 \\
& 50 & 0.115 & 0.494 & 0.655 & 0.754 & 0.901 & 0.837 \\
& 30 & 0.061 & 0.457 & 0.655 & 0.698 & 0.901 & 0.775 \\
& 15 & 0.028 & 0.428 & 0.655 & 0.654 & 0.901 & 0.726 \\
\bottomrule
\end{tabular}
\end{adjustbox}
\end{table*}

\subsection{Label Similarity Sensitivity}
\label{app:label_similarity}

We examine how different label-similarity methods affect OTQ using flat projections of COCO, COCONut, COCO-ReM, and LVIS. Table~\ref{tab:appendix_flat_projection_label_similarity} compares five methods: BERT, LQ1, OEWN, Qwen, and Strict. LQ1 assigns a similarity score of 1 to all matched labels, while Strict assigns 1 only to exact label matches. TQ and MQ remain unchanged across methods, while LQ, meanNQ, and OTQ vary. Strict yields the lowest OTQ across all four datasets, reflecting the effect of exact-match scoring on open-vocabulary labels.

\begin{table*}[t]
\caption{Flat projection results under different label-similarity methods.}
\label{tab:appendix_flat_projection_label_similarity}
\centering
\small
\setlength{\tabcolsep}{4.2pt}
\renewcommand{\arraystretch}{1.08}
\begin{adjustbox}{max width=\textwidth}
\begin{tabular}{llccccc}
\toprule
Dataset & Label sim. & OTQ$\uparrow$ & TQ$\uparrow$ & meanNQ$\uparrow$ & MQ$\uparrow$ & LQ$\uparrow$ \\
\midrule

\multirow{5}{*}{COCO}
& BERT   & 0.098 & 0.125 & 0.781 & 0.829 & 0.929 \\
& LQ1    & 0.104 & 0.125 & 0.829 & 0.829 & 0.987 \\
& OEWN   & 0.099 & 0.125 & 0.788 & 0.829 & 0.936 \\
& Qwen   & 0.095 & 0.125 & 0.761 & 0.829 & 0.903 \\
& Strict & 0.076 & 0.125 & 0.609 & 0.829 & 0.721 \\
\midrule

\multirow{5}{*}{COCONut}
& BERT   & 0.107 & 0.132 & 0.810 & 0.857 & 0.932 \\
& LQ1    & 0.113 & 0.132 & 0.857 & 0.857 & 0.987 \\
& OEWN   & 0.108 & 0.132 & 0.818 & 0.857 & 0.940 \\
& Qwen   & 0.104 & 0.132 & 0.791 & 0.857 & 0.908 \\
& Strict & 0.084 & 0.132 & 0.635 & 0.857 & 0.727 \\
\midrule

\multirow{5}{*}{COCO-ReM}
& BERT   & 0.119 & 0.140 & 0.849 & 0.904 & 0.928 \\
& LQ1    & 0.127 & 0.140 & 0.904 & 0.904 & 0.989 \\
& OEWN   & 0.120 & 0.140 & 0.857 & 0.904 & 0.936 \\
& Qwen   & 0.116 & 0.140 & 0.825 & 0.904 & 0.900 \\
& Strict & 0.092 & 0.140 & 0.652 & 0.904 & 0.707 \\
\midrule

\multirow{5}{*}{LVIS}
& BERT   & 0.115 & 0.146 & 0.764 & 0.841 & 0.882 \\
& LQ1    & 0.126 & 0.146 & 0.841 & 0.841 & 0.972 \\
& OEWN   & 0.117 & 0.146 & 0.780 & 0.841 & 0.899 \\
& Qwen   & 0.113 & 0.146 & 0.748 & 0.841 & 0.861 \\
& Strict & 0.072 & 0.146 & 0.462 & 0.841 & 0.529 \\
\bottomrule
\end{tabular}
\end{adjustbox}
\end{table*}

\clearpage
\section{Additional Benchmark Analyses}
\label{app:benchmark_analysis}

\subsection{Cross-LVLM Reference Agreement}
\label{app:pairwise_agreement}

We compare the trees generated by different LVLMs to examine whether agreement with \datasetname differs from agreement among the other models. All five tree sets use the same recursive pipeline and annotation policy. The \datasetname reference is generated using Gemini 3 Flash.

\begin{table}[h]
\caption{Pairwise OTQ agreement among LVLM-generated trees.}
\label{tab:pairwise_lvlm_agreement}
\centering
\small
\setlength{\tabcolsep}{3.5pt}
\renewcommand{\arraystretch}{1.08}
\begin{adjustbox}{max width=\linewidth}
\begin{tabular}{lccccc}
\toprule
& \datasetname & Luna & Claude & Qwen & Kimi \\
\midrule
\datasetname & -- & 0.63 & 0.61 & 0.63 & 0.69 \\
Luna & 0.63 & -- & 0.60 & 0.63 & 0.68 \\
Claude & 0.61 & 0.60 & -- & 0.62 & 0.66 \\
Qwen & 0.63 & 0.63 & 0.62 & -- & 0.69 \\
Kimi & 0.69 & 0.68 & 0.66 & 0.69 & -- \\
\bottomrule
\end{tabular}
\end{adjustbox}
\end{table}

Pairwise OTQ scores range from 0.60 to 0.69. Agreement with \datasetname falls within the range observed among the alternative LVLMs, indicating that the \datasetname reference shows a similar level of agreement with other LVLM-generated trees.

\subsection{Proposal and Grounding Analysis}
\label{app:proposal_grounding}

We examine why \datasetname contains more semantic nodes and instance masks than the alternative LVLM outputs. Table~\ref{tab:proposal_grounding} reports the number of proposed labels per image (P), the percentage successfully grounded and accepted (G), and the number of accepted semantic nodes per image (N) at each depth. D3+$\to$D4+ includes expansions from depth 3 or deeper.

\begin{table*}[h]
\caption{Proposed labels, accepted proposals, and semantic nodes by depth. Entries are P/G/N.}
\label{tab:proposal_grounding}
\centering
\small
\setlength{\tabcolsep}{3.5pt}
\renewcommand{\arraystretch}{1.08}
\begin{adjustbox}{max width=\textwidth}
\begin{tabular}{lcccc}
\toprule
LVLM & Root$\to$D1 & D1$\to$D2 & D2$\to$D3 & D3+$\to$D4+ \\
\midrule
\datasetname & 13.48 / 60.5\% / 8.16 & 20.01 / 45.8\% / 9.16 & 17.61 / 35.2\% / 6.20 & 10.33 / 15.6\% / 1.61 \\
Luna & 11.30 / 58.1\% / 6.57 & 16.03 / 47.1\% / 7.55 & 9.34 / 44.2\% / 4.13 & 3.11 / 30.3\% / 0.94 \\
Claude & 13.04 / 57.1\% / 7.44 & 17.62 / 48.8\% / 8.60 & 11.07 / 44.9\% / 4.97 & 3.90 / 30.2\% / 1.18 \\
Qwen & 10.69 / 66.5\% / 7.11 & 12.17 / 58.7\% / 7.15 & 8.15 / 49.0\% / 3.99 & 3.47 / 28.5\% / 0.99 \\
Kimi & 11.20 / 63.2\% / 7.08 & 14.26 / 53.0\% / 7.56 & 9.85 / 46.0\% / 4.53 & 4.52 / 24.5\% / 1.11 \\
\bottomrule
\end{tabular}
\end{adjustbox}

\par\smallskip

\begin{adjustbox}{max width=\textwidth}
\begin{tabular}{lrrrrr}
\toprule
\multicolumn{6}{l}{\textit{Overall}} \\
LVLM & P & G (\%) & N & Masks/img & Masks/node \\
\midrule
\datasetname & 61.42 & 40.9 & 25.12 & 83.12 & 3.31 \\
Luna & 39.77 & 48.2 & 19.19 & 62.70 & 3.27 \\
Claude & 45.62 & 48.6 & 22.19 & 65.71 & 2.96 \\
Qwen & 34.48 & 55.8 & 19.24 & 59.24 & 3.08 \\
Kimi & 39.83 & 50.9 & 20.27 & 64.72 & 3.19 \\
\bottomrule
\end{tabular}
\end{adjustbox}
\end{table*}

\datasetname proposes more labels at every depth, averaging 61.42 proposals per image compared with 34.48--45.62 for the alternative LVLMs. Although a smaller proportion of its proposals are accepted (40.9\% vs. 48.2--55.8\%), it generates more semantic nodes (25.12 vs. 19.19--22.19 per image). The number of masks per semantic node is similar across models (2.96--3.31). These results indicate that the higher annotation density of \datasetname is associated with more semantic proposals throughout recursive decomposition.

\clearpage
\subsection{Additional Baseline Results}
\label{app:baseline_details}

Table~\ref{tab:additional_baseline_results} reports additional results for the alternative LVLMs and HIPIE evaluated in Section~\ref{sec_benchmark_tree_comparison}. TP, FP, and FN denote the average numbers of matched, unmatched predicted, and unmatched reference instance nodes per image, respectively. Ext. mAR measures mask recall against four external segmentation references: COCO-17~\cite{lin2014microsoft}, COCONut~\cite{deng2024coconut}, COCO-ReM~\cite{singh2024benchmarking}, and LVIS~\cite{gupta2019lvis}.

\begin{table*}[h]
\caption{Node recovery and external mask recall results.}
\label{tab:additional_baseline_results}
\centering
\small
\setlength{\tabcolsep}{4pt}
\renewcommand{\arraystretch}{1.08}
\begin{adjustbox}{max width=\textwidth}
\begin{tabular}{lrrrr}
\toprule
Method & TP/img & FP/img & FN/img & Ext. mAR$\uparrow$ \\
\midrule
\datasetname (reference) & 83.12 & 0.00 & 0.00 & 0.57 \\
GPT-5.6 Luna & 48.58 & 14.12 & 34.54 & 0.53 \\
Claude Sonnet 5 & 47.96 & 17.75 & 35.16 & 0.53 \\
Qwen3.7 Plus & 46.05 & 13.19 & 37.07 & 0.55 \\
Kimi K3 & 52.93 & 11.80 & 30.19 & 0.55 \\
HIPIE & 10.18 & 121.92 & 72.94 & 0.36 \\
\bottomrule
\end{tabular}
\end{adjustbox}
\end{table*}

\clearpage
\section*{NeurIPS Paper Checklist}

\begin{enumerate}

\item {\bf Claims}
    \item[] Question: Do the main claims made in the abstract and introduction accurately reflect the paper's contributions and scope?
    \item[] Answer: \answerYes{}.
    \item[] Justification: The abstract and Introduction state the main contributions of the paper: \datasetname, the open tree decomposition setting, the construction pipeline, human validation, OTQ, and benchmark experiments. The claims are scoped to visible tree-structured decomposition.

\item {\bf Limitations}
    \item[] Question: Does the paper discuss the limitations of the work performed by the authors?
    \item[] Answer: \answerYes{}.
    \item[] Justification: The paper discusses limitations related to reference multiplicity, open-label evaluation, the visible-only annotation scope, automatic construction errors, and reduced reliability on deep or very small regions.

\item {\bf Theory assumptions and proofs}
    \item[] Question: For each theoretical result, does the paper provide the full set of assumptions and a complete (and correct) proof?
    \item[] Answer: \answerNA{}.
    \item[] Justification: The paper introduces a dataset, construction procedure, human validation protocol, and evaluation metric. It does not claim theoretical results or provide theorems requiring proofs.

\item {\bf Experimental result reproducibility}
    \item[] Question: Does the paper fully disclose all the information needed to reproduce the main experimental results of the paper to the extent that it affects the main claims and/or conclusions of the paper (regardless of whether the code and data are provided or not)?
    \item[] Answer: \answerYes{}.
    \item[] Justification: The paper describes the dataset construction procedure, annotation schema, human validation protocol, OTQ definition, controlled degradation analysis, and baseline setup. We release the \datasetname dataset code and data generation pipeline code needed to reproduce the main dataset artifacts. (see https://github.com/melonkick3090/COCOTree).

\item {\bf Open access to data and code}
    \item[] Question: Does the paper provide open access to the data and code, with sufficient instructions to faithfully reproduce the main experimental results, as described in supplemental material?
    \item[] Answer: \answerYes{}.
    \item[] Justification: Our dataset and code are publicly available at \projecturl.

\item {\bf Experimental setting/details}
    \item[] Question: Does the paper specify all the training and test details (e.g., data splits, hyperparameters, how they were chosen, type of optimizer) necessary to understand the results?
    \item[] Answer: \answerYes{}.
    \item[] Justification: The paper specifies the construction pipeline, validation axes, OTQ computation, controlled degradation analysis, flat mask compatibility protocol, and baseline setup. The work is primarily a dataset and benchmark, so optimizer and training hyperparameters are not central to the main claims.

\item {\bf Experiment statistical significance}
    \item[] Question: Does the paper report error bars suitably and correctly defined or other appropriate information about the statistical significance of the experiments?
    \item[] Answer: \answerYes{}.
    \item[] Justification: The main quantitative results are dataset statistics, human validation counts and percentages, metric audits, and deterministic scoring tables rather than stochastic training comparisons. The paper reports response counts and stratified validation summaries for the human evaluation.

\item {\bf Experiments compute resources}
    \item[] Question: For each experiment, does the paper provide sufficient information on the computer resources (type of compute workers, memory, time of execution) needed to reproduce the experiments?
    \item[] Answer: \answerNo{}.
    \item[] Justification: The current draft describes the construction and evaluation procedures, but complete compute worker, memory, runtime, and total compute details are not yet included.

\item {\bf Code of ethics}
    \item[] Question: Does the research conducted in the paper conform, in every respect, with the NeurIPS Code of Ethics \url{https://neurips.cc/public/EthicsGuidelines}?
    \item[] Answer: \answerYes{}.
    \item[] Justification: The work builds a vision dataset and benchmark from existing image data, includes human validation of annotation quality, and documents scope, limitations, release, and reproducibility considerations.
    \item[] Guidelines:
    \begin{itemize}
        \item The answer \answerNA{} means that the authors have not reviewed the NeurIPS Code of Ethics.
        \item If the authors answer \answerNo{}, they should explain the special circumstances that require a deviation from the Code of Ethics.
        \item The authors should make sure to preserve anonymity (e.g., if there is a special consideration due to laws or regulations in their jurisdiction).
    \end{itemize}

\item {\bf Broader impacts}
    \item[] Question: Does the paper discuss both potential positive societal impacts and negative societal impacts of the work performed?
    \item[] Answer: \answerYes{}.
    \item[] Justification: The paper discusses potential benefits of \datasetname as a dataset and benchmark for richer image understanding, including tree prediction, hierarchy training, part discovery, VLM grounding, scene analysis, and data extension. It also notes that the parent conditioned generation pipeline can be applied to other image collections, enabling open tree decomposition beyond the COCO based release.
    \item[] Guidelines:
    \begin{itemize}
        \item The answer \answerNA{} means that there is no societal impact of the work performed.
        \item If the authors answer \answerNA{} or \answerNo{}, they should explain why their work has no societal impact or why the paper does not address societal impact.
        \item Examples of negative societal impacts include potential malicious or unintended uses (e.g., disinformation, generating fake profiles, surveillance), fairness considerations (e.g., deployment of technologies that could make decisions that unfairly impact specific groups), privacy considerations, and security considerations.
        \item The conference expects that many papers will be foundational research and not tied to particular applications, let alone deployments. However, if there is a direct path to any negative applications, the authors should point it out. For example, it is legitimate to point out that an improvement in the quality of generative models could be used to generate Deepfakes for disinformation. On the other hand, it is not needed to point out that a generic algorithm for optimizing neural networks could enable people to train models that generate Deepfakes faster.
        \item The authors should consider possible harms that could arise when the technology is being used as intended and functioning correctly, harms that could arise when the technology is being used as intended but gives incorrect results, and harms following from (intentional or unintentional) misuse of the technology.
        \item If there are negative societal impacts, the authors could also discuss possible mitigation strategies (e.g., gated release of models, providing defenses in addition to attacks, mechanisms for monitoring misuse, mechanisms to monitor how a system learns from feedback over time, improving the efficiency and accessibility of ML).
    \end{itemize}

\item {\bf Safeguards}
    \item[] Question: Does the paper describe safeguards that have been put in place for responsible release of data or models that have a high risk for misuse (e.g., pre-trained language models, image generators, or scraped datasets)?
    \item[] Answer: \answerNA{}.
    \item[] Justification: The paper releases a dataset, annotations, and construction code rather than a high-risk generative model or pretrained language model. The release is based on existing vision datasets.

\item {\bf Licenses for existing assets}
    \item[] Question: Are the creators or original owners of assets (e.g., code, data, models), used in the paper, properly credited and are the license and terms of use explicitly mentioned and properly respected?
    \item[] Answer: \answerYes{}.
    \item[] Justification: The paper cites the existing datasets and methods used for construction, comparison, and positioning, including COCO, LVIS, COCONut, SA-1B, PACO, and panoptic part datasets. The release documentation state the applicable licenses and terms for reused assets.

\item {\bf New assets}
    \item[] Question: Are new assets introduced in the paper well documented and is the documentation provided alongside the assets?
    \item[] Answer: \answerYes{}.
    \item[] Justification: The paper introduces \datasetname and documents the task definition, node schema, construction pipeline, provenance, human validation protocol, OTQ metric, and benchmark setup. The released materials include the dataset annotations, dataset code, and data generation pipeline code.

\item {\bf Crowdsourcing and research with human subjects}
    \item[] Question: For crowdsourcing experiments and research with human subjects, does the paper include the full text of instructions given to participants and screenshots, if applicable, as well as details about compensation (if any)? 
    \item[] Answer: \answerYes{}.
    \item[] Justification: The paper describes the human validation axes and response summaries. Reviewer instructions, interface screenshots.

\item {\bf Institutional review board (IRB) approvals or equivalent for research with human subjects}
    \item[] Question: Does the paper describe potential risks incurred by study participants, whether such risks were disclosed to the subjects, and whether Institutional Review Board (IRB) approvals (or an equivalent approval/review based on the requirements of your country or institution) were obtained?
    \item[] Answer: \answerNA{}.
    \item[] Justification: The paper includes human validation of \datasetname annotation quality. Reviewers judge image annotations, and no sensitive personal data is collected from participants.

\item {\bf Declaration of LLM usage}
    \item[] Question: Does the paper describe the usage of LLMs if it is an important, original, or non-standard component of the core methods in this research? Note that if the LLM is used only for writing, editing, or formatting purposes and does \emph{not} impact the core methodology, scientific rigor, or originality of the research, declaration is not required.
    \item[] Answer: \answerYes{}.
    \item[] Justification: The construction section describes the use of LVLMs in the recursive parent-conditioned annotation pipeline, where they propose visible child candidates for each parent region before SAM3 grounding.

\end{enumerate}

\end{document}